%% file: main.tex
\documentclass[10pt,twocolumn,letterpaper]{article}

\usepackage[pagenumbers]{cvpr} % To force page numbers, e.g. for an arXiv version

\definecolor{cvprblue}{rgb}{0.21,0.49,0.74}
\usepackage[pagebackref,breaklinks,colorlinks,allcolors=cvprblue]{hyperref}

\def\paperID{4688} % *** Enter the Paper ID here
\def\confName{CVPR}
\def\confYear{2025}

\usepackage{natbib}

\usepackage[most]{tcolorbox}

\input{math_commands.tex}

\usepackage{hyperref}
\usepackage{url}
\usepackage{verbatim}
\usepackage{bbm}
\usepackage{amsmath}
\usepackage{booktabs}
\usepackage{xcolor}
\usepackage{amsmath, amssymb}  % for arrows
\usepackage{graphicx}
\usepackage{subcaption}
\usepackage{wrapfig}
\usepackage{placeins}
\usepackage{float}
\usepackage{todonotes}
\usepackage{xcolor}
\usepackage{tcolorbox}
\usepackage{makecell}
\usepackage{arydshln}
\usepackage{multirow}
\usepackage{mathtools}
\definecolor{myyellow}{RGB}{255, 255, 200}    % Custom yellow
\definecolor{mygray}{gray}{0.9}               % Custom light gray
\definecolor{mycyan}{RGB}{200, 255, 255}      % Custom cyan
\usepackage{amsthm}
\usepackage{bbm}
\usepackage{inconsolata}

\usepackage{listings}
\lstdefinestyle{custom}{
  basicstyle=\scriptsize\ttfamily,  % Set the font size to \footnotesize with monospaced font
  frame=single,                       % Add a frame around the listing
  breaklines=true,                    % Enable line wrapping
  columns=fullflexible,               % Allow flexibility for line breaks
  keepspaces=true,                    % Preserve spaces
  showstringspaces=false,             % Don't show spaces with underscores
  xleftmargin=0pt,                    % Remove left margin
  xrightmargin=0pt,                   % Remove right margin
  framexleftmargin=0pt,               % Remove left margin inside frame
  framexrightmargin=0pt,              % Remove right margin inside frame
  breakindent=0pt,                     % No indentation for wrapped lines
  literate={%
    {\[image\]} {{\colorbox{myyellow}{\textbf{<image>}}}}1 % Highlight <image> in red
    {\[textual\ context\]} {{\colorbox{mycyan}{\textbf{<text information>}}}}1 % Highlight <textual context> in blue
    {\[question\]} {{\colorbox{mygray}{\textbf{<question>}}}}1 % Highlight <question> in green
  }
}

\newtcolorbox{prompt}[1]{
    enhanced,
    drop shadow=black!5!white,
    left=2mm,
    right=2mm,
    top=2mm,
    bottom=2mm,
    boxsep=0mm,
    rounded corners,
    title=#1,
    colframe=black!50!white,
    fontupper=\footnotesize\linespread{0.9}\fontfamily{lmr}\selectfont,
    fontlower=\footnotesize\linespread{0.9}\fontfamily{lmr}\selectfont,
    }

\newcommand\independent{\protect\mathpalette{\protect\independenT}{\perp}}
\def\independenT#1#2{\mathrel{\rlap{$#1#2$}\mkern2mu{#1#2}}}

\makeatletter
\renewcommand{\@fnsymbol}[1]{\textdagger} % Force † symbol
\makeatother

\title{Words or Vision: Do Vision-Language Models Have Blind Faith in Text?}

\author{Ailin Deng \ \ \  Tri Cao\ \ \ \  Zhirui Chen\thanks{corresponding author}   \ \ \  Bryan Hooi\textsuperscript{\textdagger}  \\
 National University of Singapore\\
 \texttt{\{ailin,zhiruichen\}@u.nus.edu, tricao2001vn@gmail.com}\\
 \texttt{bhooi@comp.nus.edu.sg}
 }

\theoremstyle{plain}
\newtheorem{theorem}{Theorem}[section]

\newtheorem{lemma}[theorem]{Lemma}

\theoremstyle{definition}

\newtheorem{assumption}[theorem]{Assumption}
\theoremstyle{remark}
\newtheorem{remark}[theorem]{Remark}

\newtheorem*{rep@theorem}{\rep@title}
\newcommand{\newreptheorem}[2]{%
\newenvironment{rep#1}[1]{%
 \def\rep@title{\textbf{#2 \ref{##1}}}%
 \begin{rep@theorem}}%
 {\end{rep@theorem}}}
\makeatother

\newreptheorem{theorem}{Theorem}
\newreptheorem{lemma}{Lemma}

\newcommand\given[1][]{\:#1\vert\:}
\hypersetup{ 	
pdfsubject = {},
pdftitle = {Vision-Language Tug-of-war},
pdfauthor = {}
}

\begin{document} %All text i dokumentet hamnar mellan dessa taggar, allt ovanför är formatering av dokumentet
\maketitle

\begin{abstract}
Vision-Language Models (VLMs) excel in integrating visual and textual information for vision-centric tasks, but their handling of inconsistencies between modalities is underexplored. We investigate VLMs' modality preferences when faced with visual data and varied textual inputs in vision-centered settings.
By introducing textual variations to four vision-centric tasks and evaluating ten Vision-Language Models (VLMs), we discover a \emph{``blind faith in text''} phenomenon: VLMs disproportionately trust textual data over visual data when inconsistencies arise, leading to significant performance drops under corrupted text and raising safety concerns.
We analyze factors influencing this text bias, including instruction prompts, language model size, text relevance, token order, and the interplay between visual and textual certainty. While certain factors, such as scaling up the language model size, slightly mitigate text bias, others like token order can exacerbate it due to positional biases inherited from language models. To address this issue, we explore supervised fine-tuning with text augmentation and demonstrate its effectiveness in reducing text bias. Additionally, we provide a theoretical analysis suggesting that the blind faith in text phenomenon may stem from an imbalance of pure text and multi-modal data during training.
Our findings highlight the need for balanced training and careful consideration of modality interactions in VLMs to enhance their robustness and reliability in handling multi-modal data inconsistencies.
\end{abstract}
\section{Introduction}

With the rise of Vision-Language Models (VLMs) \citep{liu2023llava,dai2305instructblip,awadalla2023openflamingo}, these models are increasingly applied in complex multi-modal tasks, such as retrieval-augmented generation (RAG)~\citep{xia2024mmed} and multi-modal agents~\citep{durante2024agent,koh2024visualwebarena,kapoor2024omniact}, where they handle large, context-rich cross-modal inputs. In these practical scenarios, inconsistencies between visual and textual inputs are common, as additional textual data may be irrelevant or even misleading~\citep{shi2023large}. Despite their strong performance on vision-centric benchmarks~\citep{goyal2017making,yue2023mmmu,yu2023mm}, VLMs' capability to handle such inconsistencies remains underexplored. This gap motivates our study, as understanding and addressing VLMs’ tendencies under these conditions is essential for their safe and reliable application in real-world, multi-modal contexts.

In this work, we explore an open but underexplored question: \emph{How do VLMs handle inconsistencies between visual and textual inputs}? This question drives us to investigate the following aspects:

\begin{enumerate}
    \item \textbf{Modality Preference}: What modality do VLMs prefer when there are inconsistencies between vision and language data?
    \item \textbf{Robustness to Text Perturbation}: Can these models maintain their performance on vision-centric tasks when faced with corrupted textual data?
    \item \textbf{Influencing Factors}: What factors affect the modality preference in VLMs?
\end{enumerate}

\begin{figure}
    \centering
    \includegraphics[width=0.8\linewidth]{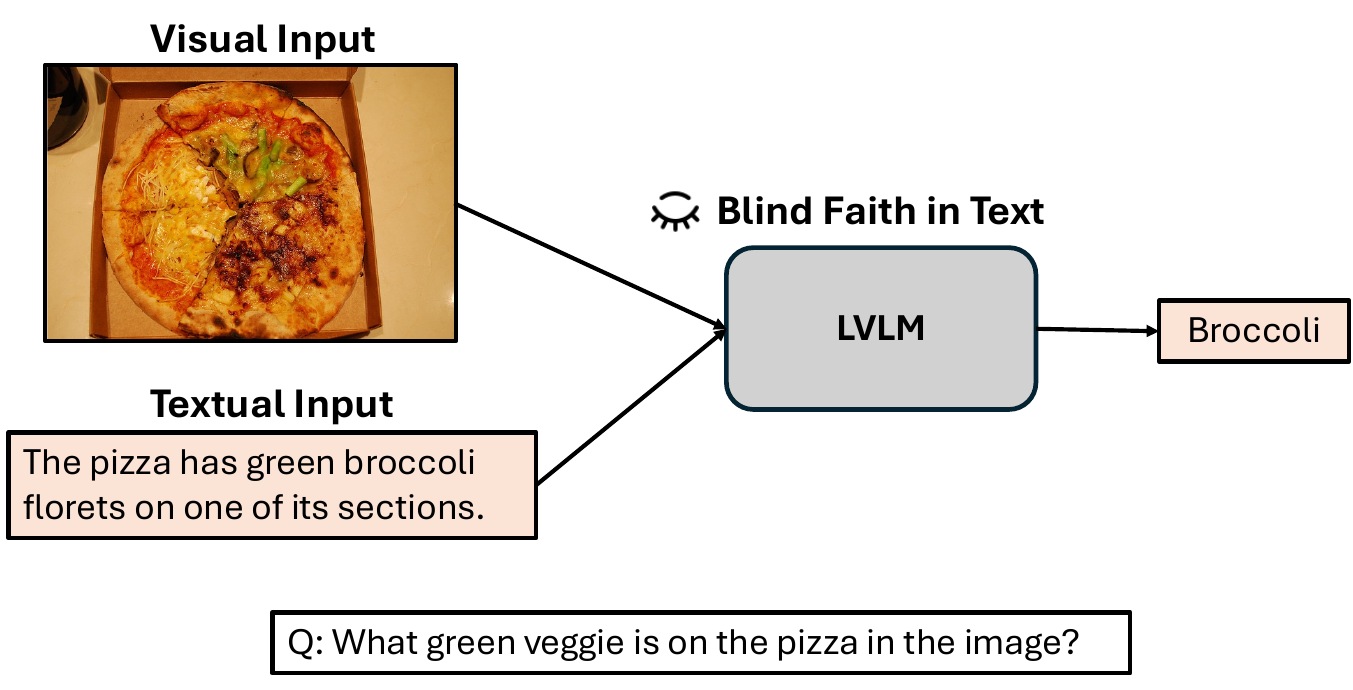}
    \caption{Illustration of the “Blind Faith in Text” phenomenon in Vision-Language Models (VLMs). These models demonstrate a strong tendency to trust textual data, when it is inconsistent with the visual data or even incorrect.}
    \label{fig:illustration}
\end{figure}

To address these questions, we construct a comprehensive benchmark by introducing textual variations to four vision-centric tasks and evaluate ten VLMs, including both proprietary and open-source models. Our findings reveal a phenomenon we term \emph{``blind faith in text''}: when inconsistencies arise between visual and textual inputs, VLMs tend to overly trust the textual data, even when it contradicts visual evidence. This text bias not only leads to significant performance degradation when the text is corrupted but also raises potential safety concerns in practical applications.

We further investigate this issue by examining factors that influence text bias:

\begin{itemize}
    \item \textbf{Instruction Prompts}: While instructions can modestly adjust modality preference, their effectiveness is limited.
    \item \textbf{Language Model Size}: Scaling up the language model size slightly mitigates text bias, but the effect saturates in larger models.
    \item \textbf{Text Relevance}: The preference for textual data increases with text relevance.
    \item \textbf{Token Order}: Placing text tokens before image tokens exacerbates text bias, possibly due to positional biases inherited from language models.
    \item \textbf{Uni-Modal Certainty}: The interplay between visual and textual certainty influences modality preference.
\end{itemize}

To mitigate text bias, we explore supervised fine-tuning with text augmentation, demonstrating its effectiveness even with limited data. Additionally, we provide a theoretical analysis suggesting that the blind faith in text phenomenon may stem from an imbalance of pure text and multi-modal data during training, as VLMs are built upon large language models primarily trained on textual data. Our contributions\footnote{\url{https://github.com/d-ailin/blind-faith-in-text}} are summarized as follows:

\begin{itemize}
    \item We uncover the \emph{``blind faith in text''} phenomenon, where VLMs prefer language data over visual data when inconsistencies occur in context.
    \item We confirm that this text bias leads to significant performance drops under text corruption, even in vision-centric tasks where VLMs typically excel.
    \item We identify key factors influencing text bias, including instruction prompts, language model size, text relevance, token order, and uni-modal certainty.
    \item We demonstrate that supervised fine-tuning with text augmentation effectively reduces text bias.
    \item We provide a theoretical analysis suggesting that the imbalance of pure text and multi-modal data during training contributes to the blind faith in text phenomenon.
\end{itemize}

\section{Preliminaries}
Given a model $f_{\rm vlm}(\cdot ; \theta)$ parameterized by $\theta$, a sample $X\coloneqq (I,T,Q)$ contains an image $I$, textual information $T$, and a question $Q$, with corresponding ground truth $Y$. We can obtain the answers under three conditions: (1) only given the image; (2) only given textual information; (3) given both modalities' information:
{\small
\begin{align}
    \hat{Y}_{\rm img} &\coloneqq f_{\rm vlm}(Q, I; \theta), \nonumber \\
    \hat{Y}_{\rm txt} &\coloneqq f_{\rm vlm}(Q, T; \theta), \nonumber \\
    \hat{Y}_{\rm mix} &\coloneqq f_{\rm vlm}(Q, I, T; \theta) \nonumber.
\end{align}
}
\paragraph{Generation Certainty.}
The response generation certainty can be estimated based on the length-normalized  
predictive likelihood of the response sequence. Formally:
{\small
\begin{align}
    \tilde{P}(\hat{Y} \mid X; \theta) \coloneqq \left( \prod_{i=1}^{\lvert \hat{Y} \rvert} \mathbb{P}(\hat{Y}_i \mid X, \hat{Y}_{<i}; \theta) \right)^{\frac{1}{\lvert \hat{Y} \rvert}}, \nonumber 
\end{align}
}where $\hat{Y}_i$ represents the $i$-th token of $\hat{Y}$, and $\hat{Y}_{<i}$ are the tokens generated before $i$-th token in $\hat{Y}$.
By considering this certainty in each modality separately, we can compute the \emph{Uni-Modal Certainty} to quantify the generation uncertainty in each modality. Specifically, it is computed given only the image or the text:
{\small
\begin{align}
    P_{\rm img} &\coloneqq \tilde{P}(\hat{Y}_{\rm img} \mid Q, I ; \theta), \nonumber \\
    P_{\rm txt} &\coloneqq \tilde{P}(\hat{Y}_{\rm txt} \mid Q, T ; \theta). \nonumber
    % P_{\textsc{i+t}} &\coloneqq L(\hat{Y}_{\textsc{i+t}} \mid Q, X_{\textsc{i}}, X_{\textsc{t}} ; \theta) \nonumber.
\end{align}
}

\subsection{Text Variations}

To comprehensively study the effect of text variations, we consider three types of variations: Match, Corruption, and Irrelevance cases given the question $Q$ and the original ground-truth answer $Y$:
\begin{itemize}
    % \item \textbf{Match.} We denote Match text cases as $(X_{\textsc{t}}^{m}, Y_{\textsc{t}}^{m})$ where $Y_{\textsc{t}}^{m} = Y$, indicating the textual information is sufficient and relevant such that ones can answer the question correctly by only relying on the given text. 
    \item \textbf{Match.} We denote $T_{\rm m}$ as a matching text such that $(T_{\rm m}, Q)$ has ground-truth $Y$, indicating the text provides sufficient and relevant information, allowing for answering the question correctly when only given the text. 
    % We take the textual information as matching when the information is correct and matching the provided image. We expect the textual information would be useful for answering the question, such as human-labeled captions for images.
    \item \textbf{Corruption.} We denote corrupted text as $T_{\rm c}$ such that $(T_{\rm c}, Q)$ has ground-truth $Y_{\rm c} \neq Y$, indicating that the text is relevant and sufficient for answering, but leads to a different answer when relying only on the text.
    \item \textbf{Irrelevance.} 
    %Note that $X=(I,T,Q)$ is a random element from some distribution.
    We consider a text to be an irrelevant text $T_{\rm irr}$ (i.e., $ T_{\rm irr} \independent I, Q$), indicating the textual information is unrelated to both the image and question, thus insufficient to answer when relying only on the text.
\end{itemize}
Starting with the base set $\mathcal{B} = \{(I,Q)\}$, we derive three variant sets by adding each text type as context: $\mathcal{Q}_{\rm m} = \{(I, T_{\rm m}, Q)\}$, $\mathcal{Q}_{\rm c} = \{(I, T_{\rm c}, Q)\}$, and $\mathcal{Q}_{\text{irr}} = \{(I, T_{\rm irr}, Q)\}$.

The motivation for these cases is to examine how models handle different text types: matching text assesses use of relevant information, corrupted text evaluates handling of misleading information, and irrelevant text checks the model’s ability to ignore distractions. Including matching text also prevents models from simply rejecting all text, as might happen if only irrelevant or corrupted text were used in prior study~\citep{wu2024clasheval}. Together, these cases provide a fuller view of VLM performance across varied text inputs.

\subsection{Model Behavior}
Given both vision and language information, we categorize the model behaviors into three conditions: (1) consistent with \textit{Image} answer ($\hat{Y}_{\rm mix} = \hat{Y}_{\rm img}$); (2) consistent with \textit{Text} answer ($\hat{Y}_{\rm mix} = \hat{Y}_{\rm txt}$); (3) \textit{Other} cases ($\hat{Y}_{\rm mix} \notin \{\hat{Y}_{\rm img}, \hat{Y}_{\rm txt}\}$). To better understand the model behavior when inconsistency between vision and language data happens, we only consider the cases where the \textit{Image} answers are different from the \textit{Text} answers (under exact match) in empirical analysis (i.e., $\hat{Y}_{\rm img} \neq \hat{Y}_{\rm txt}$). Formally, for a problem set $\mathcal{Q} \in \{\mathcal{Q}_{\rm m}, \mathcal{Q}_{\rm c}, \mathcal{Q}_{\rm irr}\}$, the proportion of the \textit{Image}, \textit{Text} and \textit{Other} answers as $p_{\rm img}$, $p_{\rm txt}$ and $p_{\rm o}$ are:

{\small
\begin{align}
    p_{\rm img} \coloneqq& \frac{ |\{X \in \mathcal{S} \mid \hat{Y}_{\rm mix} = \hat{Y}_{\rm img} \} | }{|\mathcal{S}|},\nonumber\\
    p_{\rm txt} \coloneqq& \frac{ |\{ | X \in \mathcal{S} \mid \hat{Y}_{\rm mix} = \hat{Y}_{\rm txt} \} | }{|\mathcal{S}|},\nonumber\\
    p_{\rm o} \coloneqq& \frac{ | \{X \in \mathcal{S} \mid \hat{Y}_{\rm mix} \notin \{\hat{Y}_{\rm img}, \hat{Y}_{\rm txt}\}\} |}{|\mathcal{S}|},\nonumber
\end{align}
}where $\mathcal{S} = \{X \in \mathcal{Q} \mid \hat{Y}_{\rm img} \neq \hat{Y}_{\rm txt}\}$, as we only consider the inconsistent cases where the \textit{Image} answers are different from the \textit{Text} answers.

\subsection{Metrics}

\paragraph{Text Preference Ratio (TPR).} We define TPR to quantify the model's preference for text over image-based answers. It is calculated as:
\begin{align}
\text{TPR} &\coloneqq \frac{p_{\rm txt}}{p_{\rm txt} + p_{\rm img}}. \nonumber
\end{align}
The TPR indicates text bias by showing the likelihood of the model choosing text over visual information when they are inconsistent. A higher TPR reflects a stronger text bias.
% : the models are more likely to adopt textual information beyond visual information when there are inconsistent answers with visual and textual information. 
\vspace{-1em}
\paragraph{Accuracy.} For any problem set $\mathcal{Q}$,  we have:
{\small
\begin{align}
    \text{Acc}(f_{\rm vlm}; \mathcal{Q}) = \frac{ \sum_{X \in \mathcal{Q}}{\mathbbm{1}[ f_{\rm vlm}(X) = Y]} }{| \mathcal{Q} |}.
    \nonumber
\end{align}
}
\paragraph{Macro Accuracy.} $\text{Macro}(f)$ is the average accuracy of model $f$ over different problem sets with text variations:
{\footnotesize
\begin{align}
    \text{Macro}(f_{\rm vlm}) \coloneqq \frac{1}{3} (\text{Acc}(f_{\rm vlm}; \mathcal{Q}_m) + \text{Acc}(f_{\rm vlm}; \mathcal{Q}_c) + \text{Acc}(f_{\rm vlm}; \mathcal{Q}_{\text{irr}})).
    % \text{Acc}_{\rm Macro}(f_{\rm vlm}) \coloneqq \frac{\text{Acc}(f_{\rm vlm}; \mathcal{Q}_m) + \text{Acc}(f_{\rm vlm}; \mathcal{Q}_c) + \text{Acc}(f_{\rm vlm}; \mathcal{Q}_{\text{irr}})}{3}.
    \nonumber
\end{align}
}
\paragraph{Normalized Accuracy.} $\text{Norm}(f_{\rm vlm}; \mathcal{Q})$ measures how a model is affected by the text variation, compared to the Base Accuracy, i.e., its accuracy on the base problem set $\mathcal{B}$. For any problem set $\mathcal{Q}$ under a text variation, we calculate the corresponding normalized accuracy by
{\small
\begin{align}
    \text{Norm}(f_{\rm vlm}; \mathcal{Q}) = \frac{\text{Acc}(f_{\rm vlm}; \mathcal{Q})}{\text{Acc}(f_{\rm vlm}; \mathcal{B})}.
    \nonumber
\end{align}
}

\section{Empirical Analysis}

In this section, we aim to answer the following research questions:
\begin{itemize}
    \item (\textbf{Modality Preference}) How are the models' behaviors under different text conditions? Is there any modality preference bias in the models?
    \item (\textbf{Performance Impact}) To what extent can text bias affect the models' performance, particularly with corrupted text in the context? 
    
    \item (\textbf{Influencing Factors}) Is text bias affected by instructions, size of language models, or token position?
\end{itemize}

% \vspace{-1em}
\subsection{Setup}
\label{sec:setup}
\paragraph{Tasks and Datasets.}  
We evaluate model performance on VQA datasets covering four domains, including (1) \textbf{General VQA:} 1,000 samples from VQAv2~\citep{goyal2017making} validation split; (2) \textbf{Document VQA:} 1,000 samples from 
 DocVQA~\citep{mathew2021docvqa} validation split for chart and table understanding.; 
(3) \textbf{Math Reasoning:} 1,000 samples from the minitest split of MathVista~\citep{lu2024mathvista}.
(4) \textbf{Brand Recognition:} 2,500 samples from a phishing detection dataset test split~\cite{li2024knowphish} using HTML text and webpage screenshots, focused on identifying a website’s brand.
Note, each question is expanded with three types of text variations, creating a total of 16,500 test samples derived from 5,500 unique questions. Our study includes $10$ VLMs, covering proprietary models~\citep{achiam2023gpt,anthropic2024claude3} and open models~\citep{liu2024llavanext,abdin2024phi,deitke2024molmo,wang2024qwen2}. The temperature is set to $0$ for deterministic generation. We include all experimental details, examples and results in the Appendix.

\paragraph{Text Variation Construction.}

\input{prompts/construct_text}

We use GPT-4o model to generate matching and corrupted text. Given an image $I$, question $Q$, and answer $Y$, we prompt the model to produce a supporting description as the matched text $T_m$ and a contradictory description as the corrupted text $T_c$, allowing models to produce the correct answer and an incorrect answer without image input.

The prompt used for text generation is shown in \Cref{fig:text_construct_prompt}. We extract ⟨Description 1⟩ and ⟨Description 2⟩ as the matched and corrupted texts, respectively.

To construct irrelevant text, we randomly sample passages from the WikiText dataset~\citep{merity2016pointer}, which contains texts from Wikipedia articles. These sampled texts serve as irrelevant cases, as they are factual but unrelated to the image and question. See Appendix for examples.

\paragraph{Query Instruction.}

Humans may be uncertain about which information to trust when inconsistency arises. To reduce ambiguity, following previous works~\citep{shi2023large}, we prepend a sentence to the textual information, alerting the model to potential errors and encouraging cautious use of the textual information.

{
\begin{lstlisting}[style=custom]
[image]
Here are some additional information which are text descriptions based on the image to assist you for answering the later question. Note, the information could be irrelevant, missing some information or inaccurate, please use it with caution:
[textual context]
---------------------------
[question]
\end{lstlisting}
}

\paragraph{Sanity Check.}

We evaluate the constructed text variations with model performance when only the text context is provided for answering questions. We expect that the matched text supplies enough information for correct answers, the corrupted text misleads the model into incorrect answers, and the irrelevant text lacks relevant information, leading the model to respond either randomly or with uncertainty (e.g., “I don’t know”).

\input{tables/text_only_acc}

\begin{figure}
    \centering
    \includegraphics[width=.9\linewidth]{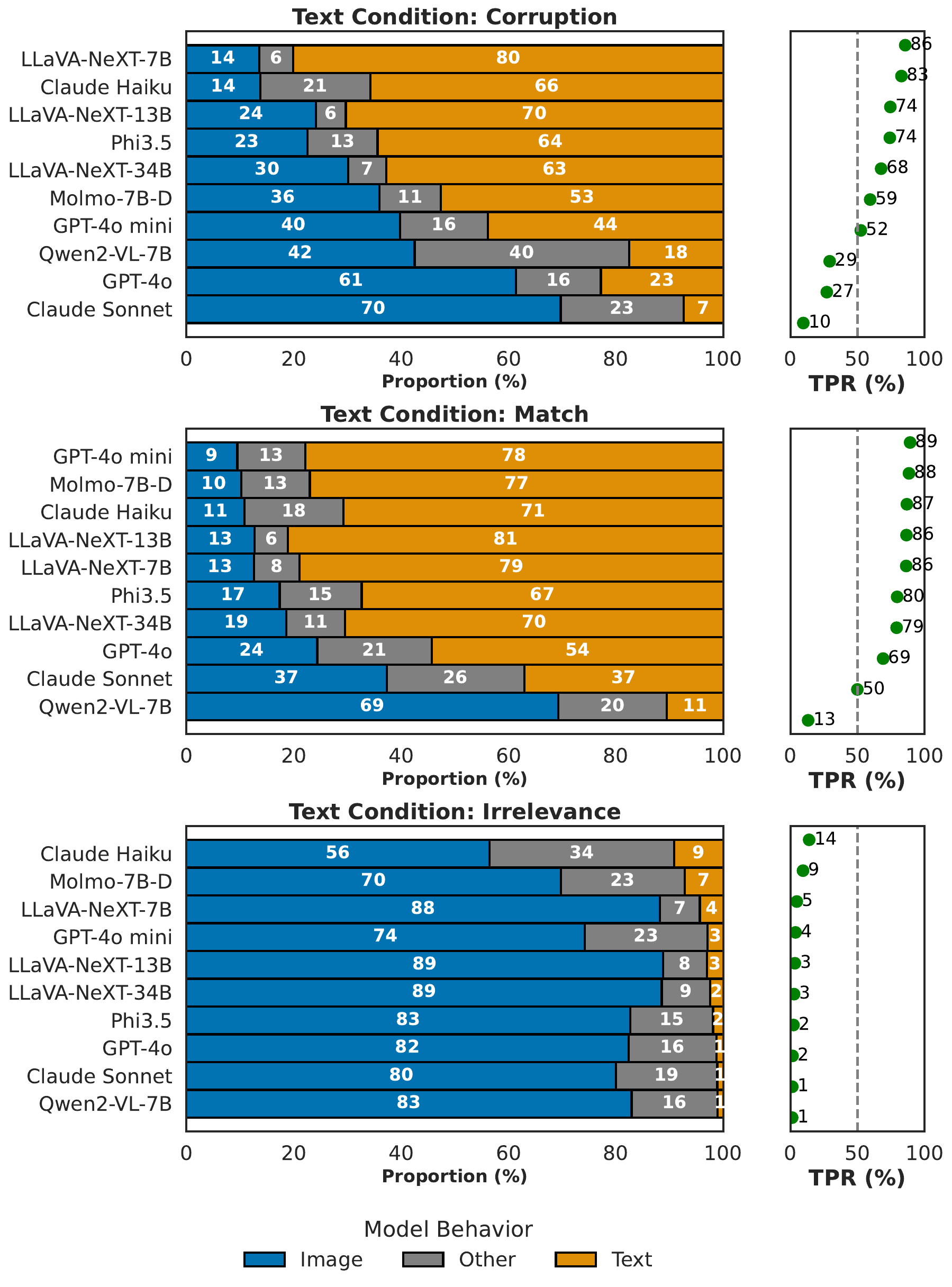}
    \caption{Model behaviors over different models when text is corrupted, matched or irrelevant.}
    \label{fig:behave_all_models}
\end{figure}

\begin{figure*}[h]
    \centering
    \includegraphics[width=.9\linewidth]{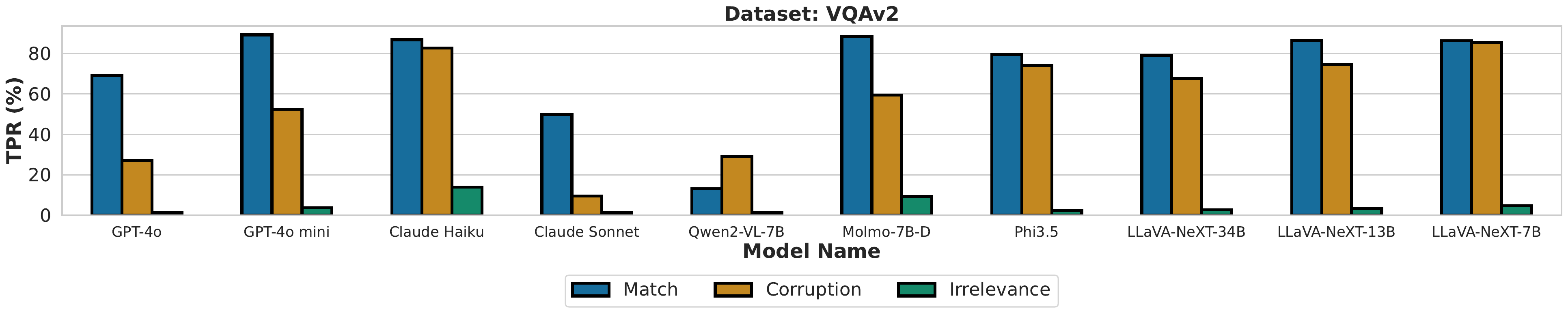}
    \vspace{-0.2cm}
    \caption{Text Preference Ratio (TPR) of all models under different text variations. Most models exhibit high text preference bias when the textual information is relevant even if they are incorrect, especially for open models. Among the proprietary models, \texttt{Claude-Sonnet} exhibits the strongest robustness to corrupted text.}
    \label{fig:text_bias_all_models}
\end{figure*}

\subsection{Blind Faith in Text}

In \Cref{fig:behave_all_models,fig:text_bias_all_models}, we observe two key findings that illustrate the phenomenon of ``blind faith in text.'' First, when textual data is inconsistent with visual data yet is relevant, models tend to favor the text, as indicated by high text preference ratios in both match and corruption cases. For example, \texttt{Claude Haiku} shows text preference ratios of $87\%$ and $83\%$ under match and corruption in VQAv2, respectively. Overall, high preference ratios are observed (usually over $50\%$), particularly for open models. Second, some models, such as \texttt{Qwen2-VL-7B}, show even higher text preference in corruption cases ($29\%$) compared to match cases ($13\%$), indicating a tendency to rely on text even when it’s incorrect, thus demonstrating limited discernment between accurate and inaccurate textual information. These results underscore the ``blind faith in text" seen across models.

\vspace{-0.5em}
\paragraph{Open models exhibit stronger text bias compared to proprietary models.}  
Although open models perform comparably to or even surpass proprietary models in standard VQA benchmarks, our results show that open models display a much higher text preference in our benchmark, even in the presence of incorrect text. This text bias remains prominent even in the efficient versions of proprietary models (i.e., \texttt{GPT-4o mini} and \texttt{Claude Haiku}). Overall, \texttt{Claude Sonnet} shows the most robustness under text-based interference among the evaluated models.
This issue is critical in the development of open models, particularly when deploying them in real-world, complex applications, such as multi-modal agents or online shopping platforms~\citep{koh2024visualwebarena,wu2024adversarial}.

\subsection{Performance Impact}  
\paragraph{The strong text bias leads to significant performance drops under corruption.}  
\input{tables/benchmark_corruption}

Given the phenomenon of ``blind faith in text," it is essential to assess its impact in performance, especially with corrupted text. As shown in \Cref{tab:benchmark_corrupt}, performance drops sharply in the presence of corrupted text. For instance, \texttt{Qwen2-VL-7B} accuracy on VQAv2, DocVQA, and MathVista falls to $59\%$, $63\%$, and $52\%$ of its original levels, an approximate $50\%$ reduction.
While proprietary models show greater stability with smaller declines, the efficient variants of these models also experience significant drops. This highlights the need for caution when deploying efficient variants of proprietary models in safety-critical applications.

\input{tables/brand_detect_corruption}

\paragraph{Text bias can introduce safety risks in real-world applications.}  
Beyond general VQA tasks, we examine the safety implications of text bias in a real-world context: brand recognition in webpage understanding~\citep{li2024knowphish}. In this task, models typically use both an HTML string and a webpage screenshot to identify a website's brand. However, HTML content can be easily manipulated with incorrect or misleading information; for example, phishing websites may inject targeted brand names into the HTML to evade detection systems, which is taken as corruption cases.
Further details about this setting are provided in the Appendix.

As shown in \Cref{tab:brand_detect_with_tpr}, under the corruption condition, most open models, such as \texttt{Molmo-7B-D}, show a significant performance drop, with accuracy reduced by nearly $50\%$ compared to the original performance. In contrast, proprietary models show slight resilience, likely due to their ability to use information from the HTML string while being less affected by injected content.

\begin{figure*}[t]
    \centering
    % First subfigure
    \begin{subfigure}[b]{.32\linewidth}
        \centering
        \includegraphics[width=\linewidth]{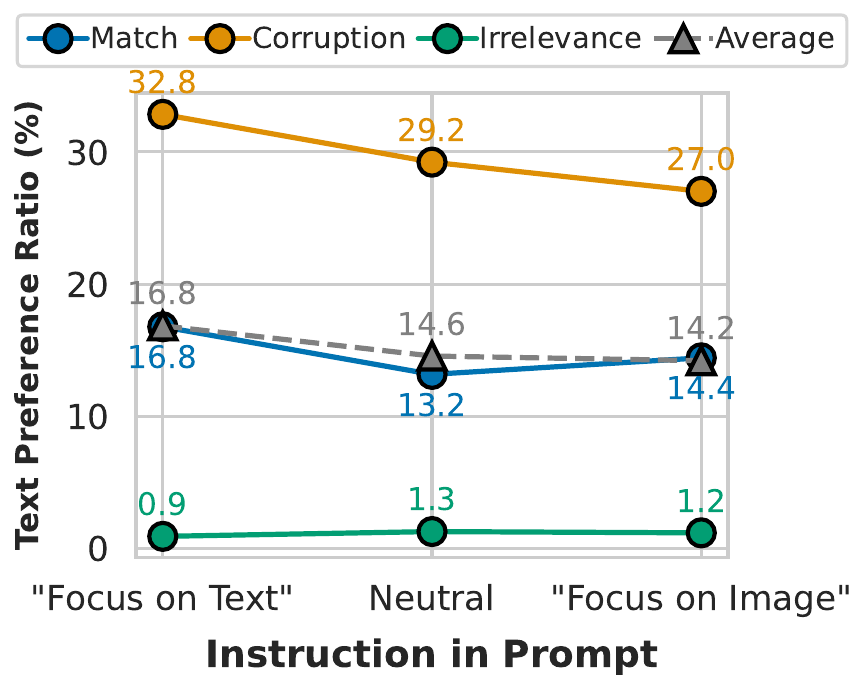} % Replace with your figure file path
        \label{fig:text_ratio_vs_prompt}
    \end{subfigure}
    % Second subfigure
    \begin{subfigure}[b]{.33\linewidth}
        \centering
        \includegraphics[width=\linewidth]{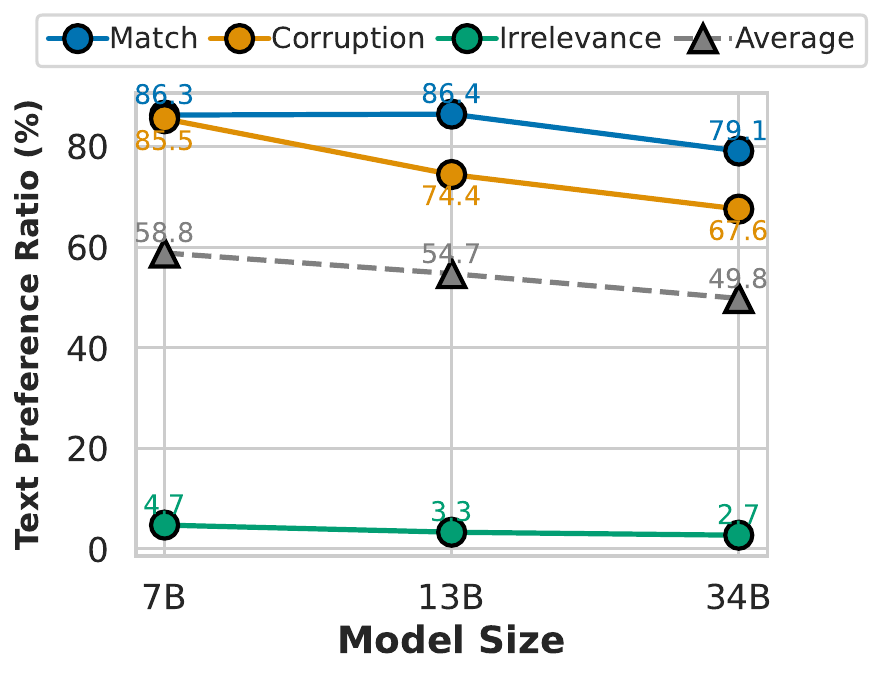} % Replace with your figure file path
        \label{fig:text_ratio_vs_size}
    \end{subfigure}
    \begin{subfigure}[b]{.29\linewidth}
        \centering
        \includegraphics[width=\linewidth]{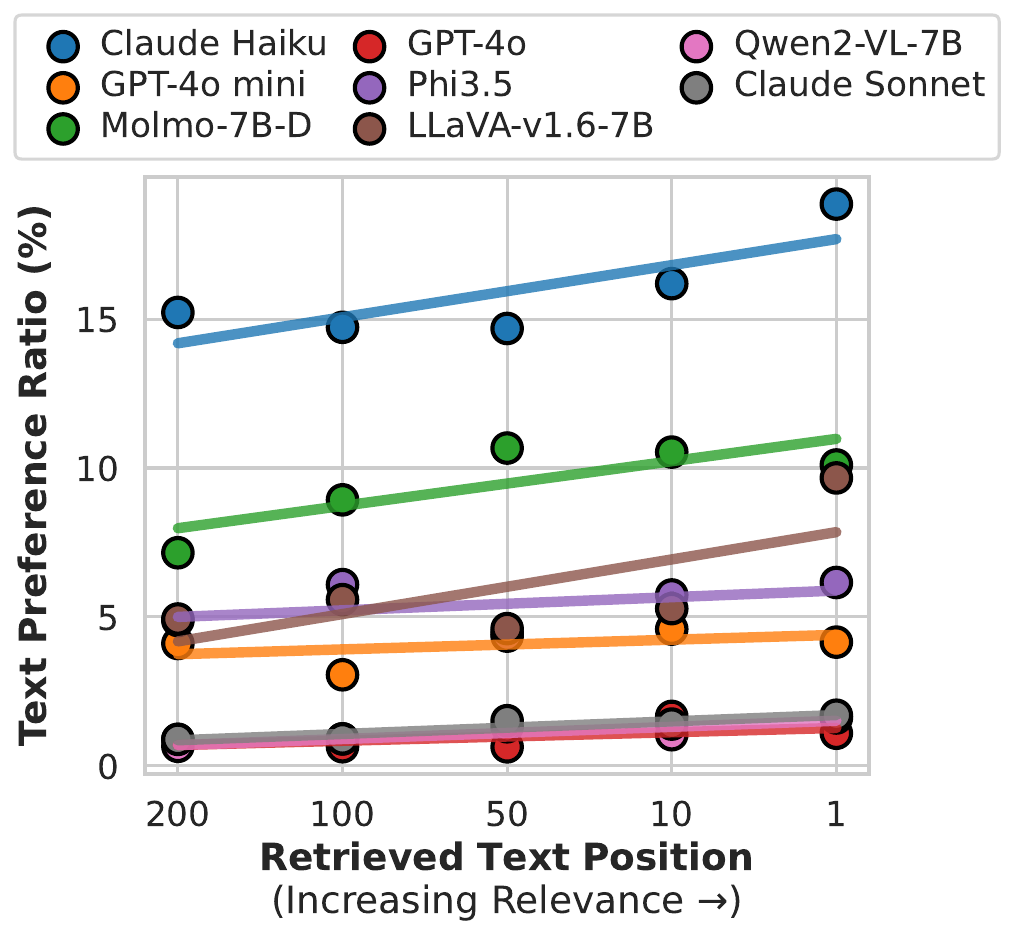} % Replace with your figure file path
        \label{fig:text_ratio_vs_rel}
    \end{subfigure}
    \vspace{-1.5em}
    % \hfill
    % Overall figure caption
    \caption{ The effect of different factors (prompting, language model size, text relevance) on text bias. \textbf{Left:} Instructional prompts influence modality preference slightly; text preference drops from $16.8\%$ to $14.2\%$ with ``Focus on Image" vs. ``Focus on Text" in \texttt{QwenVL-2-7B}. \textbf{Middle}: Scaling the language models (7B, 13B, 34B) in \texttt{LLaVA-NeXT} models decreases text bias but only marginally. \textbf{Right:} Increasing text relevance to the query with BM25 retrieval, raises text bias.}
    \label{fig:text_ratio_vs_factors}
\end{figure*}

\subsection{Influencing Factors}

In this section, we explore factors that contribute to text bias in VLMs and identify key influences. Unless otherwise noted, results are based on the VQAv2 dataset.

\vspace{-1em}
\paragraph{Instructions can reduce text bias but with limitations.}  
\label{sec:prompt_effect}  
We further investigate whether text bias can be mitigated by explicitly instructing models to focus on image information and reduce reliance on text. Inspired by previous work~\citep{shi2023large}, we prepend instructions to the questions to guide the models on which modality to prioritize. Specifically, we compare text preference ratios in three cases: neutral, ``Focus on Text," and ``Focus on Image." In the modified prompts, we add the phrases \textit{``Please focus on the text to answer the question"} and \textit{``Please focus on the image to answer the question"} respectively.
As shown in \Cref{fig:text_ratio_vs_factors} (left), the instructions influence modality preference, but the effect is limited. In \texttt{QwenVL-2-7B}, the average text preference ratio only shifts from $16.8\%$ to $14.2\%$ when changing the instruction from ``Focus on Text" to ``Focus on Image."
This limited effect may also indicate weak instruction-following capabilities in cross-modal interactions.

\paragraph{Training with a larger language model can reduce text bias but saturates.} 
Language models are essential components in current VLM architectures \citep{liu2023llava,dai2305instructblip}. Scaling up language models in VLMs generally enhances model capabilities~\citep{tong2024cambrian,karamcheti2024prismatic}. We thus study the impact of model size on text bias using the \texttt{LLaVA-NeXT} models.
As shown in \Cref{fig:text_ratio_vs_factors} (Middle), increasing model size from 7B to 34B reduces text bias overall. The 7B model exhibits high text preference with similar ratios for both match and corruption cases (86.3\% and 85.5\%, respectively). When scaled to 14B, there is a notable improvement, with a gap of $12\%$ between match and corruption text preferences. Further scaling to 34B continues to reduce text preference overall, however the gap between matched and corrupted text preference ratios remains stable.

\vspace{-0.5em}
\paragraph{Relevant text is more likely to influence vision-language models.}
In applications like RAG, retrieved text can appear relevant to a query but may ultimately be unhelpful for accurate answers. To examine how text relevance affects text bias in VLMs, we use BM25 rank retrieval \citep{robertson2009probabilistic} with the question $Q$ as the query, varying top-$k$ results to indicate relevance levels. The Top-1 result is the most relevant to the question but remains unrelated to the image, making it unhelpful for answering the question.
As shown in \Cref{fig:text_ratio_vs_factors} (Right), text bias increases with text relevance. In the most relevant (Top-1) cases, \texttt{Molmo-7B-D} exhibits over a $10\%$ text preference ratio, even though the text does not aid accurate predictions. This suggests that models are less distracted by clearly irrelevant text but are influenced by seemingly relevant (yet ultimately irrelevant) text, raising concerns for applications like multi-modal RAG, where retrieved text may appear relevant yet distract the model.

\begin{figure}[t]
    \centering
    \includegraphics[width=0.55\linewidth]{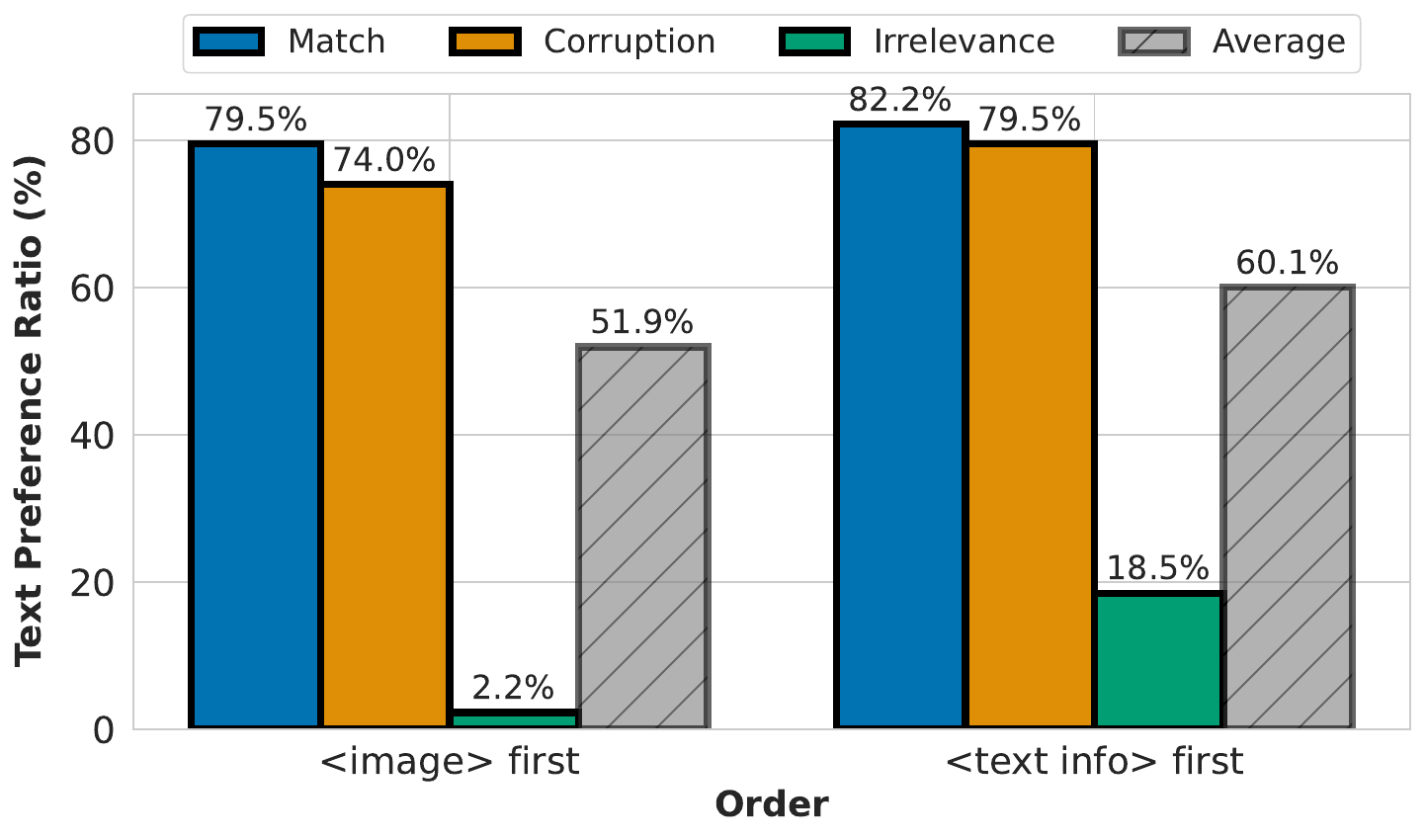}
    \vspace{-1em}
    \caption{Effect of token order on text bias: Placing text tokens before image tokens increases text bias in \texttt{Phi3.5}.}
    \label{fig:token_order}
\end{figure}

\vspace{-0.2em}
\paragraph{Text bias is related to the order of image and text tokens.} 
Previous studies have shown that token order influences bias in LLMs during language generation~\citep{zheng2023large,pezeshkpour2023large}. Since VLMs use LLMs~\citep{chiang2023vicuna,touvron2023llama2} as core components and are trained in an autoregressive manner, we examine whether text bias is affected by text and image token order. Notably, VLMs often include a large number of image tokens from vision encoder.
%(e.g., $576$ image tokens in \texttt{LLaVA-1.5}).
To test this, we compare text preference ratios by altering the order of text and image tokens in \texttt{Phi3.5}. As shown in \Cref{fig:token_order}, placing text tokens before image tokens increases text bias consistently under three text variations. While previous research has suggested that generation misalignment or hallucinations in VLMs may stem from reduced attention to image tokens \citep{zhang2024debiasing,deng2024seeing}, our findings indicate that the initial token modality may strongly influence modality preference, exacerbating text bias.

\begin{figure}[t]
    \centering
    % First subfigure
    \begin{subfigure}[b]{\linewidth}
        \centering
        \includegraphics[width=\linewidth]{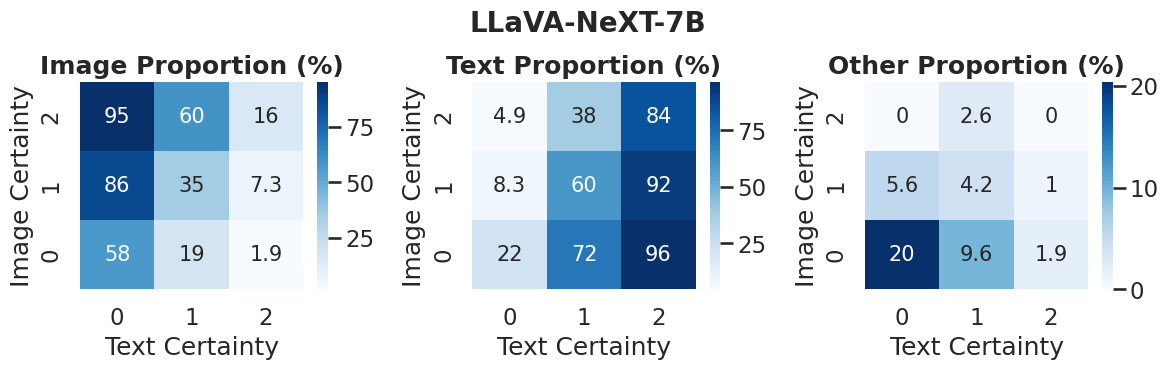} % Replace with your figure file path
    \end{subfigure}
    \hfill
    \vspace{0.1cm}
    % Second subfigure
    \begin{subfigure}[b]{\linewidth}
        \centering
        \includegraphics[width=\linewidth]{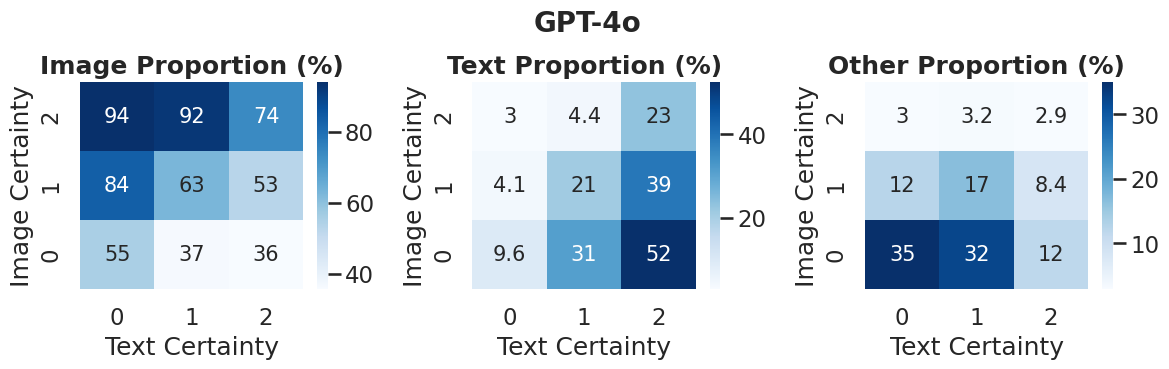} % Replace with your figure file path
    \end{subfigure}
  
    \caption{
    Effect of uni-modality certainty on model modality preference. Image/Text certainties are divided into three quantile bins, with higher values indicating higher certainty. Models favor visual data when image certainty is high and text certainty is low, and vice versa. When both certainties are low, models often produce \textit{Other} answers instead of favoring one modality alone.
    }
    \label{fig:model_behave_vs_conf}
\end{figure}

\vspace{-0.4em}
\paragraph{Interplay between uni-modal certainty and model behavior.}  
To explore when models rely on vision versus text, we explore uni-modal certainty as a key factor in shaping model behavior. Specifically, we analyze the proportions of image, text, and other responses (i.e., $p_{\rm img}$, $p_{\rm txt}$, and $p_{\rm o}$) across groups divided by uni-modal certainty quantiles. \Cref{fig:model_behave_vs_conf} shows an interesting interplay effect: when text certainty $P_{\rm txt}$ is high and image certainty $P_{\rm img}$ is low, models favor \textit{Text} answers, and vice versa. When both certainties are low, models often produce \textit{Other} answers, instead of favoring \textit{Text} or \textit{Image} alone.

\section{Investigated Solutions}

\subsection{Instruction}  
In \Cref{sec:prompt_effect}, we observed that instructional prompts can influence the model’s modality preference. For example, adding the instruction “Focus on the image to answer the question” before the question helps reduce text bias to some extent. 
To explore this further, we evaluate performance with this instruction as a baseline, finding a slight improvement ($1$–$2\%$) in Macro accuracy, as shown in \Cref{tab:ft_res,tab:ft_res_ood}.

\subsection{Supervised Finetuning (SFT)}

\paragraph{Data.}  
The composition of training data is key for effective VLM training~\citep{tong2024cambrian}. Specifically, we include both text-only and image-text samples for fine-tuning. We collect 1,000 samples evenly distributed across five data types: text-only data, original VQA data, and VQA samples under match, corruption, and irrelevance text conditions as text-augmented samples. Seed data is from the VQAv2 validation split, separate from the benchmark evaluation data.

\vspace{-.5em}
\paragraph{Setup.}  
We follow a standard supervised fine-tuning procedure, using a learning rate of $1.0 \times 10^{-4}$ with cosine decay over 3 epochs and a warmup ratio of 0.1 for stable convergence. Additionally, we apply LoRA for efficient fine-tuning. Experiments are conducted on the \texttt{LLaVA-NeXT-7B} and \texttt{Qwen2-VL-7B} models.
\vspace{-0.5em}
\paragraph{In-Distribution Performance.}  
In \Cref{tab:ft_res}, we compare the performance of the original models, models with instruction, and models after supervised fine-tuning on in-distribution data. The results show that supervised fine-tuning can better improve model accuracy compared to instruction, especially under text corruption conditions, where corruption accuracy increases from $28.69\%$ to $71.25\%$, while maintaining overall performance in macro accuracy.
\input{tables/ft_res}

\vspace{-0.5em}
\paragraph{Generalization.}  
\input{tables/ft_res_generalization}

We further assess the generalization of the fine-tuned models by evaluating their performance on datasets beyond VQAv2. As shown in \Cref{tab:ft_res_ood}, the fine-tuned models exhibit some improvement across all datasets. However, improvements are smallest on MathVista, likely due to a greater distribution shift from general VQA tasks to math reasoning tasks in vision.

\begin{figure}[t]
    \centering
    % First subfigure
    \begin{subfigure}[b]{0.54\linewidth}
        \centering
        \includegraphics[width=\linewidth]{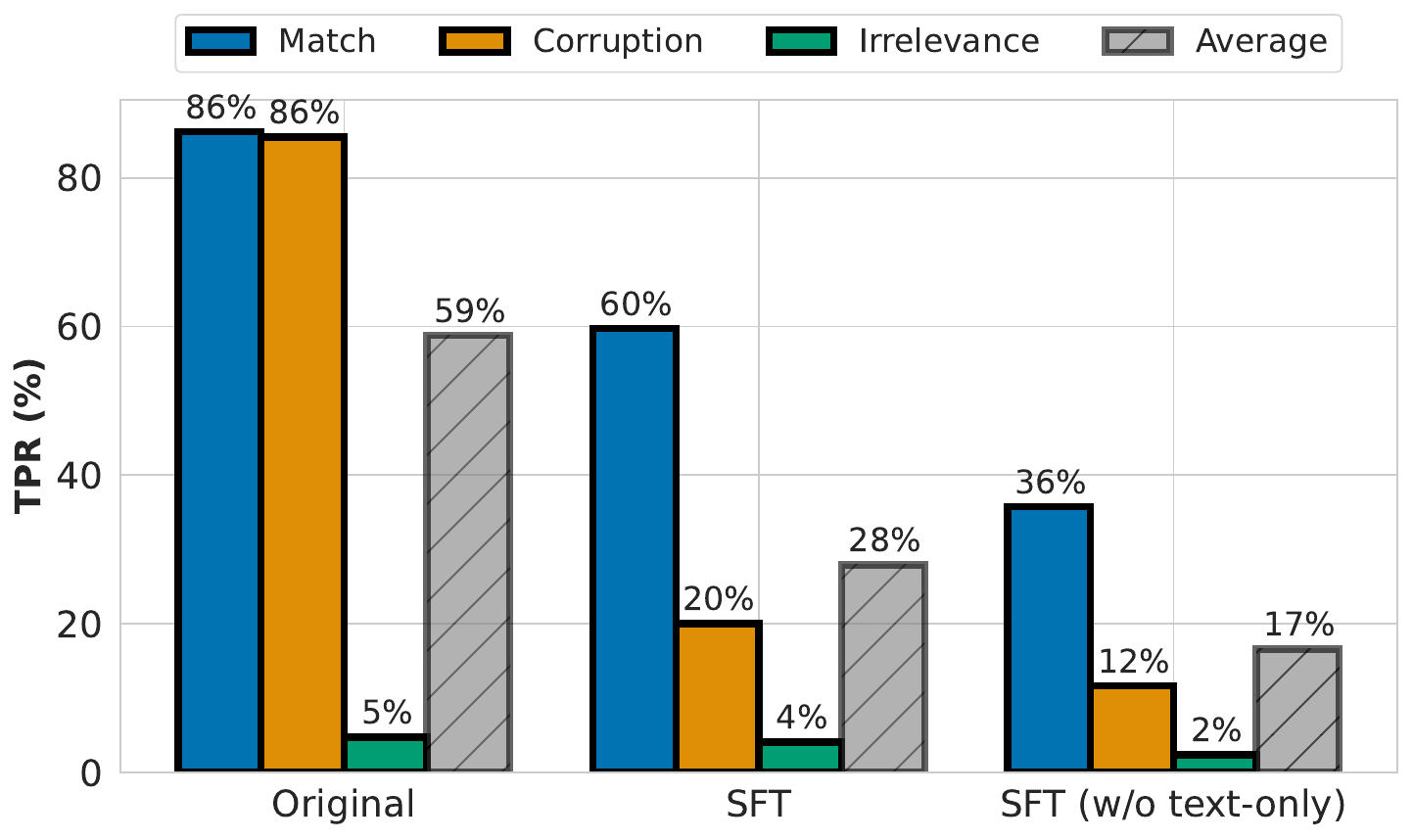} % Replace with your figure file path
    \end{subfigure}
    % Second subfigure
    \begin{subfigure}[b]{0.44\linewidth}
        \centering
        \includegraphics[width=\linewidth]{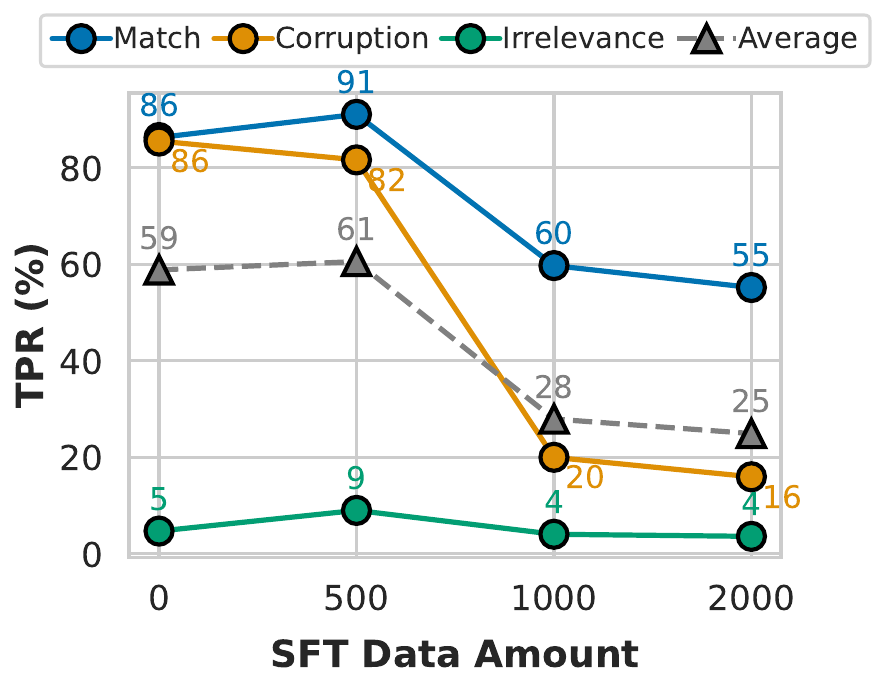} % Replace with your figure file path
    \end{subfigure}
    % Overall figure caption
    \caption{\textbf{Left:} The effect of text-only data in SFT. \textbf{Right:} The effect of data volume in SFT.}
    \label{fig:sft_data_ablation}
\end{figure}

\vspace{-.8em}
\paragraph{Effect of Text-Only Data.}  
We conduct an ablation study to inspect the role of text-only and cross-modality data in fine-tuning on \texttt{LLaVA-NeXT-7B}. For a fair comparison, the total amount of training data remains constant across experiments. As shown in \Cref{fig:sft_data_ablation} (Left), fine-tuning reduces text bias and enhances the model’s ability to distinguish between match and corruption cases, with a gap up to $40\%$. Additionally, text-only data is important for maintaining core language capabilities: without it, models may reject text indiscriminately, leading to overly cautious behavior and limiting their use of helpful text.
\vspace{-0.8em}
\paragraph{Effect of Data Volume.}  
We study the impact of data volume in SFT, shown in \Cref{fig:sft_data_ablation} (Right). As the amount of SFT data increases, the model’s reliance on text decreases significantly in corruption cases (from $58\%$ to $25\%$) while remaining relatively steady in match cases. This trend indicates that scaling up SFT data can reduce dependency on corrupted or irrelevant text, while preserving the model's effectiveness to match text.

% \paragraph{Training more steps?}

\section{Theoretical Analysis}
% \ailin{add explanation for cross-modal error}
In this section, we present theoretical analysis to explain why the majority of VLMs exhibit an inherent tendency to have blind faith in text. Let $N$ and $M$ be the size of pure-text data and multi-modal data in the training set that are i.i.d sampled from distributions $\mathcal{D}^{\rm txt}$ and $\mathcal{D}^{\rm mul}$, respectively. Our informal results are as follows, see more details in \Cref{app:detials_of_theory}.
\begin{theorem}(Informal; Theorem~\ref{thm:main} (simplified) )
Under certain assumptions, with probability at least $1-\delta$ the expected loss under pure-text data ${\mathbb{E}}_{{{(X,Y)\sim \mathcal{D}^{\rm txt}}}} \left[l(f_{\rm vlm}(X;\hat{\theta}_{\rm ERM}),Y) \right] $ achieves
{ \small
$$
 \widetilde{O} \left(\varepsilon_{\rm appr}^{\rm txt} + \frac{M}{N+M} \varepsilon_{\rm cross} + \sqrt{\frac{C_{\rm vlm}/\log(1/\delta)}{N+M}} \right),
$$
}
and similarly the expected loss under multi-modal data ${\mathbb{E}}_{{{(X,Y)\sim \mathcal{D}^{\rm mul}}}} \left[l(f_{\rm vlm}(X;\hat{\theta}_{\rm ERM}),Y) \right] $ achieves
{\small
$$
 \widetilde{O} \left(\varepsilon_{\rm appr}^{\rm mul} + \frac{N}{N+M} \varepsilon_{\rm cross} + \sqrt{\frac{C_{\rm vlm}/\log(1/\delta)}{N+M}} \right).
$$
}
\end{theorem}
\noindent $l(\cdot,\cdot)$ is a bounded loss function, and $\hat{\theta}_{\rm ERM}$ is the learned parameter(s) from Empirical Risk Minimization (ERM);  $\varepsilon_{\rm appr}^{\rm txt}$ (resp. $\varepsilon_{\rm appr}^{\rm mul}$)  and $\varepsilon_{\rm cross}$ are the quantities that represent the approximation error of pure-text data (resp. multi-modal data) and cross-modal error, respectively, and they are only dependent on the distributions $\mathcal{D}^{\rm txt}$,$\mathcal{D}^{\rm mul}$ and the hypothesis of models; $C_{\rm vlm}$ is a quantity related to the covering number of the hypothesis of models. See details in Appendix~\ref{app:detials_of_theory}. 

\begin{remark}

Observe that the expected losses under pure-text data and multi-modal data are influenced by $\frac{M}{N+M}\varepsilon_{\rm cross}$ and $\frac{N}{N+M}$ $\varepsilon_{\rm cross}$, respectively.
Our theoretical analysis, under specific assumptions, indicates that the tendency of blind faith in textual information may arise from the significant imbalance between $N$ and $M$. Particularly, in most VLMs, $N \gg M$, as these models often rely heavily on pre-trained language models, leading to the larger expected loss in multi-modal data and less in pure-text data, potentially making models favor text over image.

\end{remark}

\section{Related Work}

\vspace{-0.2em}
\paragraph{Evaluation on VLMs.}  
Current evaluation benchmarks for VLMs include single-task benchmarks~\citep{goyal2017making, mathew2021docvqa, schwenk2022okvqa, lu2024mathvista} and multi-modal benchmarks~\citep{yu2023mm, yue2023mmmu, li2023seed, liu2025mmbench} designed to assess general model capabilities across diverse tasks. Some studies also evaluate specific issues, such as hallucination~\citep{li-etal-2023-evaluating,deng2024seeing}, catastrophic forgetting~\citep{zhai2023investigating}, and robustness~\citep{yin2023survey}. However, these benchmarks are primarily vision-centric, usually treating text as question input without additional context, which limits the evaluation of models’ robustness to text variations. While text can be additional hints in specific tasks like math reasoning, current datasets~\citep{lu2024mathvista} focus on assessing reasoning skills rather than the model's ability to handle varied text inputs.
As a result, whether VLMs can reliably handle multi-modal inconsistencies remains an open question. This gap is critical for real-world applications, such as multi-modal RAG, where models encounter variable text inputs. To this end, our work studies VLM performance under different text variations, identifying a text bias that affects model reliability.

\vspace{-0.6em}
\paragraph{Benchmarks with Input Perturbation.} 
Text perturbations have been widely used in natural language tasks to evaluate model robustness and stability against distractions or misleading context~\citep{jia2017adversarial,morris2020textattack,liang2022holistic,shi2023large,xie2024adaptive,chen2024benchmarking}. In computer vision, similar efforts focus on adding imperceptible perturbations to image inputs to assess models' sensitivity to noise~\citep{goodfellow2014explaining,zhao2024evaluating}. 
Our work shifts focus from image perturbations to explore the effects of text variations on VLMs, which already excel in vision-centric benchmarks. Recent research~\citep{chen2024we} highlights data leakage in VLM benchmarks by studying performance with missing modalities. With a different goal, we investigate how VLMs manage inconsistencies between visual and textual data in vision-centered tasks, evaluating robustness in cross-modal interactions.

\section{Conclusion and Discussion}

Revisiting our core question—can VLMs reliably handle multi-modal inconsistencies?—our findings indicate that substantial challenges remain. In this work, we observe the phenomenon of \emph{“blind faith in text”} in VLMs, often relying on text over visual input when inconsistency arises, resulting in performance drops and potential safety risks.
Our analysis showed that factors like instructions, model size, text relevance, token order, and modality certainty can influence text bias. Notably, scaling model size and prompt changes alone do not resolve this issue. While supervised fine-tuning with text augmentation helps, balancing robustness and effectiveness in cross-modal settings remains challenging.
We hope this work highlights the risks of deploying VLMs in applications like multi-modal RAG, offering insights and prompting further development of more reliable and robust VLMs for cross-modal interactions.

\section*{Acknowledgement}
This research is supported by the Ministry of Education, Singapore, under the Academic Research Fund Tier 1 (FY2023) (Grant A-8001996-00-00).

{
    \small
    \bibliographystyle{ieeenat_fullname}
    \bibliography{main}
}

\appendix
\input{appendix}

\end{document}

%% file: math_commands.tex
\usepackage{amsmath,amsfonts,bm}

\def\eqref#1{equation~\ref{#1}}
\def\1{\bm{1}}

\DeclareMathAlphabet{\mathsfit}{\encodingdefault}{\sfdefault}{m}{sl}
\SetMathAlphabet{\mathsfit}{bold}{\encodingdefault}{\sfdefault}{bx}{n}

%% file: prompts/construct_text.tex
\begin{figure}[t]
    \centering
    \begin{prompt}{Text Construction}
\scriptsize
Given the image, the question and the answer, your task is to:\\
    1. Generate an accurate ⟨Description 1⟩  which can be used for answering the question correctly without using the image.\\
    2. Generate a wrong description ⟨Description 2⟩  which can be used for answering the question with a completely wrong answer ⟨answer 2⟩  without using the image.\\
    3. Make sure both descriptions are sound and concise.\\
    4. The wrong description's sentence structure should be similar to the correct description.\\
    
    Here are the questions and answers:\\
    Question: \{question\}\\
    Answer: \{answer\}\\
    \\
    Please output the two statements in this format:\\
    Description 1: ⟨Description 1⟩\\
    Description 2: ⟨Description 2⟩ \\
    Answer 2: ⟨answer 2⟩
\end{prompt}
    \vspace{-0.8em}
    \caption{Prompt for generating matched and corrupted text given an image, the question and the ground-truth answer. We substitute \texttt{\{question\}} and \texttt{\{answer\}} with the specific sample.}
    \label{fig:text_construct_prompt}
\end{figure}

%% file: tables/text_only_acc.tex
\begin{table}[htbp]
    \centering
    \small
    % \setlength{\tabcolsep}{8pt}
    % \renewcommand{\arraystretch}{1.0}
    % \resizebox{0.8\textwidth}{!}{
    \scalebox{0.9}{
    \begin{tabular}{l|cccc}
        \hline
        & \multicolumn{4}{c}{\textbf{VQAv2}} \\
       
        Model & Base & Match & Corruption & Irrelevance \\
        \hline
        Claude Sonnet & 66.88 & 84.39 & 16.17 & 24.39 \\
        GPT-4o & 78.39 & 90.07 & 17.59 & 18.67 \\
      
        Molmo-7B-D & 76.33 & 88.98 & 18.74 & 35.40 \\
       
        \hline
    \end{tabular}}
    \caption{Text-only accuracy (\%) across different models. It provides a sanity check for the constructed text when matched, corrupted, or irrelevant.}
    \label{tab:text_only_acc}
\end{table}

%% file: tables/benchmark_corruption.tex
\begin{table*}[htbp]
\centering
\setlength{\tabcolsep}{4pt}
% \renewcommand{\arraystretch}{1.0}
% \resizebox{0.8\textwidth}{!}{
\footnotesize
\begin{tabular}{l|cccc|cccc|cccc}
\hline
& \multicolumn{4}{c|}{ \textbf{VQAv2}} & \multicolumn{4}{c|}{ \textbf{DocVQA}} & \multicolumn{4}{c}{\textbf{MathVista}} \\
Model & Base $\uparrow$ & Corruption $\uparrow$ & Norm $\uparrow$ & TPR $\downarrow$ & 
Base $\uparrow$ & Corruption $\uparrow$ & Norm $\uparrow$ & TPR $\downarrow$ & 
Base $\uparrow$ & Corruption $\uparrow$ & Norm $\uparrow$ & TPR $\downarrow$ \\
\hline

    GPT-4o mini & 69.82 & 51.55 & 73.83 & 52.42 & 69.40 & 38.20 & 55.04 & 52.07 & 52.30 & 23.90 & 45.70 & 80.28 \\
    Claude Haiku & 50.08 & 25.54 & 50.99 & 82.70 & 68.80 & 40.20 & 58.43 & 47.67 & 41.00 & 19.80 & 48.29 & 77.42 \\
    GPT-4o & 78.39 & \textbf{70.75} & \underline{90.25} & \underline{27.09} & 85.00 & \underline{73.60} & \underline{86.59} & 
\underline{17.96} & \textbf{58.90} & \underline{41.20} & 69.95 & \underline{48.98} \\
    Claude Sonnet & 66.88 & \underline{68.17} & \textbf{101.93} & \textbf{9.58} & \underline{87.00} & \textbf{84.60} & \textbf{97.24} & \textbf{3.21} & \underline{56.30} & \textbf{49.30} & \textbf{87.57} & \textbf{29.14} \\

    \hdashline[0.5pt/2pt]
    LLaVA-NeXT-7B & 79.45 & 28.69 & 36.10 & 85.52 & 53.60 & 10.00 & 18.60 & 87.77 & 35.80 & 19.70 & 54.97 & 84.19 \\
    LLaVA-NeXT-13B & 81.02 & 37.61 & 46.40 & 74.43 & 57.70 & 11.00 & 19.10 & 86.84 & 36.20 & 20.60 & 56.89 & 80.83 \\
    LLaVA-NeXT-34B &  \underline{82.96} & 42.87 & 51.70 & 67.56 & 64.00 & 15.10 & 23.61 & 82.69 & 34.00 & 21.70 & 61.98 & 67.64 \\ %83.21
    Phi3.5 & 75.65 & 35.23 & 46.50 & 74.05 & 78.20 & 50.50 & 64.60 & 40.51 & 43.10 & 22.20 & 51.47 & 80.20 \\
    Molmo-7B-D & 76.33 & 49.29 & 64.50 & 59.40 & 74.00 & 38.40 & 51.90 & 57.20 & 44.90 & 32.90 & \underline{73.27} & 60.63 \\
    Qwen2-VL-7B & \textbf{85.51} & 50.79 & 59.41 & 29.22 & \textbf{90.50} & 57.50 & 63.63 & 37.41 & 
    55.40 & 28.90 & 52.18 & 70.23\\
    % \hline
    % Average & 74.95 & 43.93 & 59.05 & 60.15 & 63.22 & 35.53 & 48.88 & 58.41 & 42.90 & 26.57 & 61.7 & 70.34 \\
\hline
\end{tabular}
\vspace{-0.2cm}
\caption{Performance (\%) reported as Base Accuracy, Corruption Accuracy, Normalized Corruption Accuracy (Norm) and Text Preference Ratio (TPR) under corruption. \textbf{Bold}: best performance; \underline{underline}: second best. Full results under all text variations are in the Appendix.}
\label{tab:benchmark_corrupt}
\end{table*}

%% file: tables/brand_detect_corruption.tex
\begin{table}[htbp]
    \centering
    \footnotesize
    \setlength{\tabcolsep}{4pt}
    \begin{tabular}{l|cccc}
        \hline
        & \multicolumn{4}{c}{\textbf{Brand Detection}} \\
        Model & Base $\uparrow$ & Corruption $\uparrow$ & Norm $\uparrow$ & TPR $\downarrow$ \\
        \hline
        GPT-4o mini & 88.84 & 84.8 & 95.44 & 7.48 \\
        Claude Haiku & 84.40 & 78.72 & 93.27 & 6.44 \\
        GPT-4o & 88.68 & \textbf{89.76} & \underline{101.22} & \textbf{0.83} \\
        Claude Sonnet & \textbf{90.20} & \underline{90.24} & 100.04 & \underline{0.96} \\
        \hdashline[0.5pt/2pt]
       
        LLaVA-NeXT-7B & 78.60 & 55.32 & 70.39 & 59.17 \\
        LLaVA-NeXT-13B & 83.00 & 60.00 & 72.29 & 40.65 \\
        LLaVA-NeXT-34B & 66.28 & 53.52 & 80.77 & 23.49\\
        Phi3.5 & 84.40 & 60.68 & 71.90 & 50.45 \\
        Molmo-7B-D & 87.44 & 41.44 & 47.39 & 60.40 \\
        Qwen2-VL-7B & \underline{89.68} & 86.48 & 96.43 & 2.99\\
        \hline
    \end{tabular}
    \caption{Performance on the Brand Detection dataset reported in Base Accuracy, Corruption Accuracy, Normalized Corruption Accuracy (Norm), and Text Preference Ratio (TPR). \textbf{Bold}: best performance; \underline{underline}: second best performance.}
    \label{tab:brand_detect_with_tpr}
\end{table}

%% file: tables/ft_res.tex
\begin{table}[t]
    \centering
    % \small
    \setlength{\tabcolsep}{2pt}
    % \renewcommand{\arraystretch}{1.0}
    % \resizebox{0.8\textwidth}{!}{
    \footnotesize
    % \scalebox{0.95}{
    \begin{tabular}{l|ccccc}
        \hline
        & \multicolumn{5}{c}{\textbf{VQAv2}} \\
       
        Model & Base $\uparrow$ & Match $\uparrow$ & Corruption $\uparrow$ & Irrelevance $\uparrow$ & Macro $\uparrow$ \\
        \hline
        LLaVA-NeXT-7B & \textbf{79.45} & \textbf{92.32} & 28.69 & \textbf{79.43} & 66.81 \\
        Instruction & \textbf{79.45} & 92.25 & 34.27 & 78.15 & 68.22 \\
        SFT & 77.48 & 87.56 & \textbf{71.25} & 77.32 & \textbf{78.71 }\\
        
        \hline
        Qwen2-VL-7B & \textbf{85.51} & \textbf{92.76} & 50.79 & 83.70 & 75.75 \\
        Instruction & \textbf{85.51} & 92.62 & 54.78 & 82.82 & 76.74 \\
        SFT & 84.18 & 87.01 & \textbf{82.72} & \textbf{84.00} & \textbf{84.58} \\
        
        \hline
    \end{tabular}
    % }
    \caption{In-distribution performance comparison between original models, instruction and fine-tuned models.}
    \label{tab:ft_res}
\end{table}

%% file: tables/ft_res_generalization.tex
\begin{table}[t]
\centering
\setlength{\tabcolsep}{2.5pt}
% \renewcommand{\arraystretch}{1.0}
% \resizebox{0.99\linewidth}{!}{
\footnotesize
\begin{tabular}{l|cc|cc|cc}
\hline
 & \multicolumn{2}{c|}{ \textbf{DocVQA}} & \multicolumn{2}{c|}{\textbf{MathVista}} & \multicolumn{2}{c}{\textbf{Brand Detection}} \\
Model & Base $\uparrow$ & Macro $\uparrow$ &  Base $\uparrow$ & Macro $\uparrow$ & 
 Base $\uparrow$ & Macro $\uparrow$\\

\hline
% \hline
LLaVA-NeXT-7B & \textbf{53.60} & 51.07 & \textbf{35.80} & 41.03 & 78.60 & 46.44 \\
Instruction & \textbf{53.60} & 49.27 & \textbf{35.80} & 41.20 & 78.60 & 47.36 \\
SFT & 52.20 & \textbf{56.17} & 35.30 &\textbf{41.63} & \textbf{81.36} & \textbf{72.29} \\
\hline
Qwen2-VL-7B & \textbf{90.50} & 80.83 & 55.40 & 53.87 & \textbf{89.68} & 81.85 \\
Instruction & \textbf{90.50} & 80.77 & 55.40 & 54.10 & \textbf{89.68} & 84.48 \\
SFT & 90.30 & \textbf{88.97} & \textbf{58.50} & \textbf{57.17} & 89.44 & \textbf{88.75} \\
\hline
\end{tabular}
%}
\caption{Performance comparison with Base and Macro accuracy based on DocVQA, MathVista, and Brand Recognition. See full results under different text conditions in Appendix.}
\label{tab:ft_res_ood}
\end{table}

%% file: appendix.tex
% \onecolumn
\clearpage
\input{theory}

\section{Experimental Setup}

This section outlines the experimental setup, including examples of constructed textual variations, details of the brand detection task~\citep{li2024knowphish}, and the evaluation protocols employed. We present examples illustrating the three types of textual variations alongside the corresponding image, original question, and ground-truth answers to provide clarity and context.

\subsection{Examples}
This subsection provides examples of matching, corrupted, and irrelevant texts across different datasets in \Cref{table:example_vqav2,table:example_docvqa,table:example_mathvista,table:example_brandreg}.
\input{tables/app_examples}

\subsection{Brand Recognition}

Brand recognition from a webpage is a crucial step in detecting phishing websites. Phishing webpages aim to deceive users by imitating the appearance of legitimate websites associated with well-known brands. Accurately identifying the brand linked to a webpage allows for a comparison between the input webpage's URL and the official URL of the recognized brand, aiding in the detection of phishing attempts.

In our experiments, we utilized phishing webpage samples from the TR-OP dataset \cite{li2024knowphish}. Each sample comprises a screenshot and its corresponding HTML code. Depending on the scenario, the HTML content either reflects the target brand displayed in the screenshot or is altered to assess the model's robustness. We evaluated three specific scenarios:

\begin{itemize}
    \item \textbf{Matching:} The original HTML includes information about the target brand visible in the screenshot. This scenario provides the model with consistent inputs, helping it correctly identify the brand.
    \item \textbf{Corruption:} In this case, we inserted a fabricated brand name (e.g., ``The official webpage of MobrisPremier'') into the HTML to mislead the model into recognizing a non-existent brand. Since no corresponding URL exists for such brands, phishing detection becomes infeasible for these inputs.
    \item \textbf{Irrelevance:} The HTML content was replaced with randomly selected sentences from the Wiki dataset \cite{}, ensuring that the new content was unrelated to any brand. This scenario tests the model's ability to handle inputs with no brand-specific information.
\end{itemize}

To standardize the inputs, we preprocessed the HTML content by removing all tags and truncating it to a maximum length of 5,000 characters.

\subsection{Evaluation}

We follow the evaluation protocol specified for each dataset. To reduce cases where models generate open-ended answers, which complicates evaluation, we adopt a similar approach to the evaluation setting in \texttt{LLaVA-1.5} \citep{liu2024improved}. For certain datasets, we append additional formatting prompts after the question, as shown in \Cref{tab:format_prompt}.

\begin{table}[h]
    \centering
    \begin{tabular}{l|l}
        \hline
        \textbf{Dataset} & \textbf{Response Formatting Prompts} \\
        \hline
        VQAv2 \citep{goyal2017making} & Please only output the answer with a single word or phrase. \\
        DocVQA \citep{mathew2021docvqa} & Please only output the answer directly. \\
        MathVista \citep{lu2024mathvista} & -- \\
        Brand Recognition \citep{li2024knowphish} & Only output the brand name without any additional information. \\
        \hline
    \end{tabular}
    \caption{Response formatting prompts used for evaluation.}
    \label{tab:format_prompt}
\end{table}

For MathVista \citep{lu2024mathvista}, which uses GPT-based evaluation, we do not include formatting prompts. Instead, GPT is employed directly to evaluate the outputs.

\section{Experimental Results}

To rigorously assess the performance impact of varying textual contexts, we have documented the comprehensive results across four distinct datasets. These results are quantified using several metrics: Accuracy, Normalized Accuracy, and Text Preference Ratio (TPR) for the text variations of Match, Corruption, and Irrelevance, alongside Macro Accuracy. The detailed outcomes are encapsulated in Table \ref{app:table:full_res}.

\input{tables/app_full_res}

For a thorough assessment of the investigated methodologies, encompassing base models, instructional prompts, and Supervised Fine-Tuning (SFT), we present results across four datasets, measured in terms of Accuracy, Normalized Accuracy, Text Preference Ratio (TPR) under the text variations of Match, Corruption, and Irrelevance, as well as Macro Accuracy. These experiments were conducted utilizing the models \texttt{LLaVA-NeXT-7B} and \texttt{Qwen2-VL-7B}. The detailed findings are provided in Table \ref{app:tab:ft_res}.

\input{tables/app_ft_res_generalization}

%% file: theory.tex
\onecolumn
\section{Details of Theoretical Analysis}
\label{app:detials_of_theory}
To provide a rigorous foundation for our theoretical analysis, we begin by formally outlining the training process of a vision-language model. For clarity and conciseness, the following is a streamlined adaptation of the standard training process. A VLM is a function $f_{\rm vlm} : \mathcal{X} \rightarrow \mathcal{Y}$, where $\mathcal{X} \coloneqq \mathbb{R}^{\tau \times d}$ denotes the set of sequences of $d$-dimensional feature vector (that can represent text or image) with length $\tau$, and $\mathcal{Y}$ denotes the output space of the model. Without loss of generalization, we assume $\mathcal{Y} \coloneqq \mathbb{R}$ for simplicity.

\subsection{Structure}
Following \citet{edelman2022inductive}, we consider the form of transformer structure of $f_{\rm vlm}$ with $L$ layers as follows. The parameters of $i$'s layer is denoted by $W^{(i)}:=\left\{W_Q^{(i)}, W_K^{(i)}, W_V^{(i)}, W_C^{(i)}\right\}$. In addition, we denote $W^{1: i}=\left(W^{(1)}, \ldots, W^{i-1}\right)$ to be the parameters up to $i$'s layer. Further, we let the block of $i$-th layer $g_{\text {tf-block }}^{(i)}: \mathbb{R}^{\tau \times d} \rightarrow \mathbb{R}^{\tau \times d}$ to be
\begin{align}
& g_{\text {tf-block }}^{(i+1)}\left(X ; W^{1: i+1}\right):= \Pi_{\text {norm }}\left(\sigma\left(\Pi_{\text {norm }}\left(f\left(X\right)\right)\right) W_C^{(i)}\right) \text{ for } i=1, \nonumber \\
& g_{\text {tf-block }}^{(i+1)}\left(X ; W^{1: i+1}\right):= \Pi_{\text {norm }}\left(\sigma\left(\Pi_{\text {norm }}\left(f\left(g_{\text {tf-block }}^{(i)}\left(X ; W^{1: i}\right) ; W^{(i)}\right)\right)\right) W_C^{(i)}\right) \text{ for } i>1, \nonumber 
\end{align}
where $X\in \mathbb{R}^{\tau \times d}$ is the model's input,  and $\Pi_{\text {norm }}$ is the layer normalization function, $\sigma$ is a non-linear activation function, and
\begin{equation*}
 f\left(Z ;\left\{W_Q, W_K, W_V\right\}\right):=  \text { Softmax }\left(Z W_Q\left(Z W_K\right)^{\top}\right) Z W_V
\end{equation*}
with ${\rm Softmax}(\cdot)$ being the standard softmax function. Finally, the scalar output is defined as 
\begin{equation}
\label{eq:vlm}
f_{\rm vlm}( X; W^{1: L}, w) \coloneqq w^\top [g_{\text {tf-block }}^{(L+1)}\left(X ; W^{1: L}\right)]_\tau, \text{ for some } w \in \mathbb{R}^d,
\end{equation}
where $[\mathbf{G}]_\tau \in \mathbb{R}^d$ denotes the $\tau$-th row of the matrix $\mathbf{G} \in \mathbb{R}^{\tau \times d}$. 
Furthermore, we have the following assumptions within the structure.
\begin{assumption}
\label{asp:1}
For all $i=1,\cdots,L$, we have $\left\|W_V^{(i)}\right\|_2,\left\|W_K^{(i)} W_Q^{(i)^{\top}}\right\|_2,\left\|W_C^{(i)}\right\|_2 \leq C_2$.
\end{assumption}

\begin{assumption}
\label{asp:2}
For all $i=1,\cdots,L$, we have $\left\|W_V^{(i)}\right\|_{2,1},\left\|W_K^{(i)^{\top}} W_Q^{(i)}\right\|_{2,1},\left\|W_C^{(i)}\right\|_{2,1} \leq C_{2,1}$.
\end{assumption}

\begin{assumption}
\label{asp:3}
The activation function $\sigma(\cdot)$ is $L_\sigma$-Lipschitz in the $l_2$ norm.
\end{assumption}

\begin{assumption}
\label{asp:4}
The loss function $l(\cdot)$ is $b$-bounded and is $L_{\rm loss}$-Lipschitz in its arguments.
\end{assumption}

\subsection{Training process}
\noindent
Let $\mathcal{X}^{\rm txt} = [ (X^{\rm txt}_1,y_1^{\rm txt}),\cdots, (X^{\rm txt}_N,y_N^{\rm txt} ) ]$ be a pure-text training set with size $N$, where $X^{\rm txt}_i \in \mathbb{R}^{\tau\times d}$ is a sequence of the text feature vector of length $\tau$, and $y_i^{\rm txt}=f_{\rm gt}^{\rm txt} (X^{\rm txt}_i ) \in \mathbb{R}$ is its ground-truth label with $f_{\rm gt}^{\rm txt}(\cdot ) $ denoted as the ground-true function for the pure text data. We assume  $X^{\rm txt}_1,\cdots, X^{\rm txt}_N $ are i.i.d. sampled from a unknown distribution $\mathcal{D}^{\rm txt}$.

In addition, let $\mathcal{X}^{\rm mul} = [ (X^{\rm mul}_1,y_1^{\rm mul}),\cdots, (X^{\rm mul}_N,y_M^{\rm mul} ) ]$ be a multi-modal training set with size $M$, where $X^{\rm multi}_i \in \mathbb{R}^{\tau\times d}$ is a sequence of multi-modal (e.g., text and image) feature vector of length $\tau$, and $y_i^{\rm mul}=f_{\rm gt}^{\rm mul} (X^{\rm multi}_i ) \in \mathbb{R}$ is its ground-truth label with $f_{\rm gt}^{\rm mul}(\cdot ) $ denoted as the ground-true function for the multi-modal data. Similarly, we assume  $X^{\rm mul}_1,\cdots, X^{\rm mul}_N$ are i.i.d. sampled from a unknown distribution $\mathcal{D}^{\rm mul}$.

Furthermore, let $l : \mathbb{R}\times \mathbb{R} \rightarrow $ be a loss function. Then, we define the parameter $\hat{\theta}_{\rm ERM} \in \Theta$ according to the ERM learning process of the multi-modal paradigm as 
\begin{equation}
\hat{\theta}_{\rm ERM} \in \arg \min_{\theta \in \Theta} \frac{1}{N+M} \left(\sum_{i=1}^N l\left(f_{\rm vlm}( X_i^{\rm txt}; \theta), y^{\rm txt}_i \right) + \sum_{i=1}^M l\left(f_{\rm vlm}( X_i^{\rm mul}; \theta), y^{\rm mul}_i \right) \right)
\end{equation}
Our main theoretical result is given in the next subsection.

\subsection{Results}
We now provide the formal statement of Theorem~\ref{thm:main}. 
\begin{theorem}
\label{thm:main}
Let $\Theta$ be the set of parameters that satisfies Assumption~\ref{asp:1},~\ref{asp:2},~\ref{asp:3} and~\ref{asp:4}. For any $\theta \in \Theta$, let  $f_{\rm vlm}(\cdot; \theta)$ be a VLM as is defined in~\eqref{eq:vlm} with $L$ layers. With probability at least $1-\delta$, 
{\small	
\begin{flalign} 
      & \underset{{{X} \sim \mathcal{D}^{\rm txt}} }{\mathbb{E}} \big[ l \left(f_{\rm vlm}(X; \hat{\theta}_{\rm ERM}), f^{\rm txt}_{\rm gt}(X)  \right) \big] \nonumber \\
      & \lesssim  \underbrace{\inf_{\theta \in \Theta}  \underset{{{X} \sim \mathcal{D}^{\rm txt}} }{\mathbb{E}} \big[ l \left(f_{\rm vlm}(X; \theta, f^{\rm txt}_{\rm gt}(X)  \right) \big]}_\text{\footnotesize  approximation error} + \underbrace{\frac{M}{M+N} \sup_{\theta \in \Theta} \left|\underset{{{X} \sim \mathcal{D}^{\rm mul}} }{\mathbb{E}} \big[ l \left(f_{\rm vlm}(X; {\theta}), f^{\rm mul}_{\rm gt}(X)  \right) \big]  - \underset{{{X} \sim \mathcal{D}^{\rm txt}} }{\mathbb{E}} \big[ l \left(f_{\rm vlm}(X; {\theta}), f^{\rm txt}_{\rm gt}(X)  \right) \big]  \right|}_\text{\footnotesize  cross-modal error}  \nonumber \\
     & \hspace{.2in} + \underbrace{b \sqrt{\frac{1/\log(\delta)}{N+M}} +  L_{
      \rm loss
      } \cdot \sqrt{\frac{C_{\rm vlm}}{N+M}} \cdot \log(1+\frac{N+M}{C_{\rm vlm}})}_\text{\footnotesize generalization error},
\end{flalign}
}
and 
{\small	
\begin{flalign} 
      & \underset{{{X} \sim \mathcal{D}^{\rm mul}} }{\mathbb{E}} \big[ l \left(f_{\rm vlm}(X; \hat{\theta}_{\rm ERM}), f^{\rm mul}_{\rm gt}(X)  \right) \big] \nonumber \\
      & \lesssim  \underbrace{\inf_{\theta \in \Theta}  \underset{{{X} \sim \mathcal{D}^{\rm mul}} }{\mathbb{E}} \big[ l \left(f_{\rm vlm}(X; \theta, f^{\rm mul}_{\rm gt}(X)  \right) \big]}_\text{\footnotesize approximation error} + \underbrace{\frac{N}{M+N} \sup_{\theta \in \Theta} \left|\underset{{{X} \sim \mathcal{D}^{\rm mul}} }{\mathbb{E}} \big[ l \left(f_{\rm vlm}(X; {\theta}), f^{\rm mul}_{\rm gt}(X)  \right) \big]  - \underset{{{X} \sim \mathcal{D}^{\rm txt}} }{\mathbb{E}} \big[ l \left(f_{\rm vlm}(X; {\theta}), f^{\rm txt}_{\rm gt}(X)  \right) \big]  \right|}_\text{\footnotesize  cross-modal error}  \nonumber \\
      & \hspace{.2in} + \underbrace{b \sqrt{\frac{1/\log(\delta)}{N+M}} +  L_{
      \rm loss
      } \cdot \sqrt{\frac{C_{\rm vlm}}{N+M}} \cdot \log(1+\frac{N+M}{C_{\rm vlm}})}_\text{\footnotesize generalization error},
\end{flalign}
}
where $$C_{\rm vlm} \lesssim \left(C_2 L_\sigma\right)^{O(L)} \cdot {B_X^2 B_w^2 C_{2,1}^2} \cdot \log (d \tau (N+M)) $$ 
is the constant related to the covering number of the function class of $\{f_{\rm vlm}(\cdot; \theta) \given \theta \in \Theta \}$,
and the notation $\lesssim$ hides global constants and logarithmic factors on quantities besides $N,M$ and $\tau$.

\end{theorem}

\subsection{Proof of Theorem~\ref{thm:main}}
Before we formally prove Theorem~\ref{thm:main}, we first present some useful Lemmas from previous works. For any real-valued function class $\mathcal{F}$, we let $ \mathcal{N}_{\infty}\left(\mathcal{F} ; \varepsilon ; x^{(1)}, \ldots, x^{(m)}\right)$ denote the converting number of  $\mathcal{F}$ with respect to 
the radius $\varepsilon$ and the samples $\{x^{(1)}, \ldots, x^{(m)} \}$.
 \begin{lemma}(Adapted from \citet[Theorem 8]{bartlett2002rademacher} and \citet[Lemma A.4]{edelman2022inductive})
 \label{lemma:1}
 Consider a real-valued function class $\mathcal{F}$ such that $|f| \leq A$ for all $f \in \mathcal{F}$ and $\log \mathcal{N}_{\infty}\left(\mathcal{F} ; \varepsilon ; x^{(1)}, \ldots, x^{(m)}\right) \leq C_{\mathcal{F}} / \varepsilon^2$ for all $x^{(1)}, \ldots, x^{(m)} \in \mathcal{X}$. Let $l(\cdot,\cdot)$ to be a loss function bounded by $b$ and is $L_{\rm loss}$-Lipschitz in its arguments, and $g_{\rm gt}:\mathcal{X}\rightarrow \mathbb{R}$ be a ground-true function.  Then for any $\delta>0$ and any distribution $\mathcal{D}$ for the i.i.d samples  $x^{(1)}, \ldots, x^{(m)} \in \mathcal{X}$, with probability at least $1-\delta$, simultaneously for all $f \in \mathcal{F}$,
$$
\left| \underset{x \sim \mathcal{D}}{\mathbb{E}} \left[l(f(x),g_{\rm gt}(x)) \right] -\frac{1}{m}\sum_{i=1}^m l\left(f(x^{(i)}),g_{\rm gt}(x^{(i)})\right) \right| \leq 4 c L_{\rm loss} \sqrt{\frac{C_{\mathcal{F}}}{m}}\left(1+\log \left(A \sqrt{m / C_{\mathcal{F}}}\right)\right)+2 b \sqrt{\frac{\log (1 / \delta)}{2 m}}
$$
for some constant $c>0$.
\end{lemma}

\begin{lemma}(Adapted from \citet[Theorem A.17]{edelman2022inductive})
 \label{lemma:2}
 Suppose  $\forall i \in[m],\left\|X^{(i)}\right\|_{2, \infty} \leq B_X $\text. Let $\Theta$ be the set of parameters that satisfies Assumption~\ref{asp:1},~\ref{asp:2},~\ref{asp:3} and~\ref{asp:4}. For any $\theta \in \Theta$, let  $f_{\rm vlm}(\cdot; \theta)$ is a vlm model as is fined in ~\eqref{eq:vlm} with $L$ layers. We have
$$
\log \mathcal{N}_{\infty}\left( \{ f_{\rm vlm}(\cdot; \theta) \given \theta \in \Theta \} ; \varepsilon ; X^{(1)}, \ldots, X^{(m)}\right) \lesssim\left(C_2 L_\sigma\right)^{O(L)} \cdot \frac{B_X^2 B_w^2 C_{2,1}^2}{\varepsilon^2} \cdot \log (d m T).
$$
\end{lemma}

\begin{proof}[Proof of Theorem~\ref{thm:main}]
By Lemma~\ref{lemma:1},  with probability at least $1-\delta$ we have simultaneously for all $\theta \in \Theta$,
\begin{align}
 & \left| \underset{{{X} \sim \mathcal{D}^{\rm txt}} }{\mathbb{E}} \big[ l \left(f_{\rm vlm}(X; {\theta}), f^{\rm txt}_{\rm gt}(X)  \right) \big] 
 -  \frac{1}{N} \sum_{i=1}^N l \left(f_{\rm vlm}(X^{\rm txt}_i); {\theta}), f^{\rm txt}_{\rm gt}(X^{\rm txt}_i)  \right) \right| \nonumber \\
  & \le 4 c L_{\rm loss} \sqrt{\frac{C}{N}}\left(1+\log \left(A \sqrt{N / C}\right)\right)+2 b \sqrt{\frac{\log (1 / \delta)}{2 N}}, \label{eqn:comm1}
\end{align}
where $A\le (C_2 L_\sigma)^{2L} \cdot B_X$, and $C$ is a constant such that for all $\varepsilon>0$ and $X^{(1)}, \ldots, X^{(m)} \in \mathbb{R}^{\tau\times d}$ with $\lVert X^{(i)} \rVert_{2,\infty} \le B_X$
$$
\log \mathcal{N}_{\infty}\left( \{ f_{\rm vlm}(\cdot; \theta) \given \theta \in \Theta \} ; \varepsilon ; X^{(1)}, \ldots, X^{(m)}\right)  \le \frac{C}{\varepsilon^2}.
$$
Similarly, with probability at least $1-\delta$ we have simultaneously for all $\theta \in \Theta$,
\begin{align}
 & \left| \underset{{{X} \sim \mathcal{D}^{\rm mul}} }{\mathbb{E}} \big[ l \left(f_{\rm vlm}(X; {\theta}), f^{\rm mul}_{\rm gt}(X)  \right) \big] 
 -  \frac{1}{M} \sum_{i=1}^M l \left(f_{\rm vlm}(X^{\rm mul}_i); {\theta}), f^{\rm mul}_{\rm gt}(X^{\rm mul}_i)  \right) \right| \nonumber \\
  & \le 4 c L_{\rm loss} \sqrt{\frac{C}{M}}\left(1+\log \left(A \sqrt{M / C}\right)\right)+2 b \sqrt{\frac{\log (1 / \delta)}{2 M}}. \label{eq:gen_error_mul}
\end{align}
Note that for any $\theta \in \Theta$ we have
\begin{align}
& \left| \frac{1}{N+M} \left(\sum_{i=1}^N l\left(f_{\rm vlm}( X_i^{\rm txt}; \theta), y^{\rm txt}_i \right) + \sum_{i=1}^M l\left(f_{\rm vlm}( X_i^{\rm mul}; \theta), y^{\rm mul}_i \right) \right) - \underset{{{X} \sim \mathcal{D}^{\rm txt}} }{\mathbb{E}} \big[ l \left(f_{\rm vlm}(X; {\theta}), f^{\rm txt}_{\rm gt}(X)  \right) \big] 
  \right| \nonumber \\
&=  \bigg| \frac{N}{M+M} \left( \frac{1}{N} \sum_{i=1}^N l\left(f_{\rm vlm}( X_i^{\rm txt}; \theta), y^{\rm txt}_i \right) -  \underset{{{X} \sim \mathcal{D}^{\rm txt}} }{\mathbb{E}} \big[ l \left(f_{\rm vlm}(X; {\theta}), f^{\rm txt}_{\rm gt}(X)  \right) \big] \right) \nonumber \\
&\hspace{.3in} +\frac{M}{M+N} \left( \frac{1}{M} \sum_{i=1}^M l\left(f_{\rm vlm}( X_i^{\rm mul}; \theta), y^{\rm mul}_i \right) -  \underset{{{X} \sim \mathcal{D}^{\rm txt}} }{\mathbb{E}} \big[ l \left(f_{\rm vlm}(X; {\theta}), f^{\rm txt}_{\rm gt}(X)  \right) \big] \right) \bigg| \\
&\stackrel{(a)}{\le}  \frac{N}{M+M} \left| \frac{1}{N} \sum_{i=1}^N l\left(f_{\rm vlm}( X_i^{\rm txt}; \theta), y^{\rm txt}_i \right) -  \underset{{{X} \sim \mathcal{D}^{\rm txt}} }{\mathbb{E}} \big[ l \left(f_{\rm vlm}(X; {\theta}), f^{\rm txt}_{\rm gt}(X)  \right) \big] \right| \nonumber \\
&\hspace{.3in} +\frac{M}{M+N} \left| \frac{1}{M} \sum_{i=1}^M l\left(f_{\rm vlm}( X_i^{\rm mul}; \theta), y^{\rm mul}_i \right) -  \underset{{{X} \sim \mathcal{D}^{\rm txt}} }{\mathbb{E}} \big[ l \left(f_{\rm vlm}(X; {\theta}), f^{\rm txt}_{\rm gt}(X)  \right) \big] \right|, \label{eqn:comm2}
\end{align}
where (a) follows from Jensen's inequality.

In addition, with probability at least $1-\delta$,we have  for all $\theta \in \Theta$
\begin{align}
& \left| \frac{1}{M} \sum_{i=1}^M l\left(f_{\rm vlm}( X_i^{\rm mul}; \theta), y^{\rm mul}_i \right) -  \underset{{{X} \sim \mathcal{D}^{\rm txt}} }{\mathbb{E}} \big[ l \left(f_{\rm vlm}(X; {\theta}), f^{\rm txt}_{\rm gt}(X)  \right) \big] \right| \nonumber \\
& \le \left| \frac{1}{M} \sum_{i=1}^M l\left(f_{\rm vlm}( X_i^{\rm mul}; \theta), y^{\rm mul}_i \right) - \underset{{{X} \sim \mathcal{D}^{\rm mul}} }{\mathbb{E}} \big[ l \left(f_{\rm vlm}(X; {\theta}), f^{\rm mul}_{\rm gt}(X)  \right) \big] \right|  \nonumber \\
& \hspace{.3in} + \left|  \underset{{{X} \sim \mathcal{D}^{\rm txt}} }{\mathbb{E}} \big[ l \left(f_{\rm vlm}(X; {\theta}), f^{\rm txt}_{\rm gt}(X)  \right) \big] -\underset{{{X} \sim \mathcal{D}^{\rm mul}} }{\mathbb{E}} \big[ l \left(f_{\rm vlm}(X; {\theta}), f^{\rm mul}_{\rm gt}(X)  \right) \big] \right| \nonumber  \\
&\stackrel{(a)}{\le}   4 c L_{\rm loss} \sqrt{\frac{C}{M}}\left(1+\log \left(A \sqrt{M / C}\right)\right)+2 b \sqrt{\frac{\log (1 / \delta)}{2 M}}  \nonumber \\
& \hspace{.3in} + \left|  \underset{{{X} \sim \mathcal{D}^{\rm txt}} }{\mathbb{E}} \big[ l \left(f_{\rm vlm}(X; {\theta}), f^{\rm txt}_{\rm gt}(X)  \right) \big] -\underset{{{X} \sim \mathcal{D}^{\rm mul}} }{\mathbb{E}} \big[ l \left(f_{\rm vlm}(X; {\theta}), f^{\rm mul}_{\rm gt}(X)  \right) \big] \right| \label{eq:comm3} 
\end{align}
where (a) follows from~\eqref{eq:gen_error_mul}.

Combining~\eqref{eqn:comm1},~\eqref{eqn:comm2} and~\eqref{eq:comm3}, we get that with probability at least $1-\delta$, for all $\theta \in \Theta$,
\begin{align}
& \left| \frac{1}{N+M} \left(\sum_{i=1}^N l\left(f_{\rm vlm}( X_i^{\rm txt}; \theta), y^{\rm txt}_i \right) + \sum_{i=1}^M l\left(f_{\rm vlm}( X_i^{\rm mul}; \theta), y^{\rm mul}_i \right) \right) - \underset{{{X} \sim \mathcal{D}^{\rm txt}} }{\mathbb{E}} \big[ l \left(f_{\rm vlm}(X; {\theta}), f^{\rm txt}_{\rm gt}(X)  \right) \big] 
  \right| \nonumber \\
&{\le}   4 c L_{\rm loss} \frac{\sqrt{{MC}}}{N+M}\left(1+\log \left(A \sqrt{M / C}\right)\right)+2 b \frac{\sqrt{{M\log (1 / \delta)/2}}}{N+M} \nonumber \\ 
& \hspace{.3in} + 4 c L_{\rm loss} \frac{\sqrt{{NC}}}{N+M}\left(1+\log \left(A \sqrt{N / C}\right)\right)+2 b \frac{\sqrt{{N\log (1 / \delta)/2}}}{N+M} \nonumber \\
&  \hspace{.3in} + \left|  \underset{{{X} \sim \mathcal{D}^{\rm txt}} }{\mathbb{E}} \big[ l \left(f_{\rm vlm}(X; {\theta}), f^{\rm txt}_{\rm gt}(X)  \right) \big] -\underset{{{X} \sim \mathcal{D}^{\rm mul}} }{\mathbb{E}} \big[ l \left(f_{\rm vlm}(X; {\theta}), f^{\rm mul}_{\rm gt}(X)  \right) \big] \right| \label{eq:comm4}.
\end{align}

By the definition of $\hat{\theta}_{\rm ERM}$,~\eqref{eq:comm4} implies that with probability at least $1-\delta$,
{ \footnotesize	
\begin{align}
& \left| \frac{1}{N+M} \left(\sum_{i=1}^N l\left(f_{\rm vlm}( X_i^{\rm txt}; \hat{\theta}_{\rm ERM}), y^{\rm txt}_i \right) + \sum_{i=1}^M l\left(f_{\rm vlm}( X_i^{\rm mul}; \hat{\theta}_{\rm ERM}), y^{\rm mul}_i \right) \right) - \underset{{{X} \sim \mathcal{D}^{\rm txt}} }{\mathbb{E}} \big[ l \left(f_{\rm vlm}(X; {\hat{\theta}_{\rm ERM}}), f^{\rm txt}_{\rm gt}(X)  \right) \big] 
  \right| \nonumber \\
& \le \inf_{\theta \in \Theta}  \underset{{{X} \sim \mathcal{D}^{\rm mul}} }{\mathbb{E}} \big[ l \left(f_{\rm vlm}(X; \theta, f^{\rm mul}_{\rm gt}(X)  \right) \big] + \frac{N}{M+N} \sup_{\theta \in \Theta} \left|\underset{{{X} \sim \mathcal{D}^{\rm mul}} }{\mathbb{E}} \big[ l \left(f_{\rm vlm}(X; {\theta}), f^{\rm mul}_{\rm gt}(X)  \right) \big]  - \underset{{{X} \sim \mathcal{D}^{\rm txt}} }{\mathbb{E}} \big[ l \left(f_{\rm vlm}(X; {\theta}), f^{\rm txt}_{\rm gt}(X)  \right) \big]  \right| \nonumber \\
&\hspace{.2in} +   4 c L_{\rm loss} \frac{\sqrt{{MC}}}{N+M}\left(1+\log \left(A \sqrt{M / C}\right)\right)+2 b \frac{\sqrt{{M\log (1 / \delta)/2}}}{N+M}  + 4 c L_{\rm loss} \frac{\sqrt{{NC}}}{N+M}\left(1+\log \left(A \sqrt{N / C}\right)\right)+2 b \frac{\sqrt{{N\log (1 / \delta)/2}}}{N+M} \nonumber \\
&  \hspace{.2in} + \left|  \underset{{{X} \sim \mathcal{D}^{\rm txt}} }{\mathbb{E}} \big[ l \left(f_{\rm vlm}(X; {\hat{\theta}_{\rm ERM}}), f^{\rm txt}_{\rm gt}(X)  \right) \big] -\underset{{{X} \sim \mathcal{D}^{\rm mul}} }{\mathbb{E}} \big[ l \left(f_{\rm vlm}(X; {\hat{\theta}_{\rm ERM}}), f^{\rm mul}_{\rm gt}(X)  \right) \big] \right| \label{eq:comm5}.
\end{align}
}
Note the fact that $\max\{N,M\} \le N+M \le 2\max\{N,M\}$  . Finally, by Lemma~\ref{lemma:2} and  hiding global constants and logarithmic factors on quantities besides $N,M$ and $\tau$, we get with probability $1-\delta$,
{\small	
\begin{flalign} 
      & \underset{{{X} \sim \mathcal{D}^{\rm txt}} }{\mathbb{E}} \big[ l \left(f_{\rm vlm}(X; \hat{\theta}_{\rm ERM}), f^{\rm txt}_{\rm gt}(X)  \right) \big] \nonumber \\
      & \lesssim  \inf_{\theta \in \Theta}  \underset{{{X} \sim \mathcal{D}^{\rm txt}} }{\mathbb{E}} \big[ l \left(f_{\rm vlm}(X; \theta, f^{\rm txt}_{\rm gt}(X)  \right) \big] + \frac{M}{M+N} \sup_{\theta \in \Theta} \left|\underset{{{X} \sim \mathcal{D}^{\rm mul}} }{\mathbb{E}} \big[ l \left(f_{\rm vlm}(X; {\theta}), f^{\rm mul}_{\rm gt}(X)  \right) \big]  - \underset{{{X} \sim \mathcal{D}^{\rm txt}} }{\mathbb{E}} \big[ l \left(f_{\rm vlm}(X; {\theta}), f^{\rm txt}_{\rm gt}(X)  \right) \big]  \right|  \nonumber \\
      & \hspace{.2in} + b \sqrt{\frac{1/\log(\delta)}{N+M}} +  L_{
      \rm loss
      } \cdot \sqrt{\frac{C_{\rm vlm}}{N+M}} \cdot \log(1+\frac{N+M}{C_{\rm vlm}}),
\end{flalign}
}
and similarly,
{\small	
\begin{flalign} 
      & \underset{{{X} \sim \mathcal{D}^{\rm mul}} }{\mathbb{E}} \big[ l \left(f_{\rm vlm}(X; \hat{\theta}_{\rm ERM}), f^{\rm mul}_{\rm gt}(X)  \right) \big] \nonumber \\
      & \lesssim  \inf_{\theta \in \Theta}  \underset{{{X} \sim \mathcal{D}^{\rm mul}} }{\mathbb{E}} \big[ l \left(f_{\rm vlm}(X; \theta, f^{\rm mul}_{\rm gt}(X)  \right) \big] + \frac{N}{M+N} \sup_{\theta \in \Theta} \left|\underset{{{X} \sim \mathcal{D}^{\rm mul}} }{\mathbb{E}} \big[ l \left(f_{\rm vlm}(X; {\theta}), f^{\rm mul}_{\rm gt}(X)  \right) \big]  - \underset{{{X} \sim \mathcal{D}^{\rm txt}} }{\mathbb{E}} \big[ l \left(f_{\rm vlm}(X; {\theta}), f^{\rm txt}_{\rm gt}(X)  \right) \big]  \right|  \nonumber \\
      & \hspace{.2in} + b \sqrt{\frac{1/\log(\delta)}{N+M}} +  L_{
      \rm loss
      } \cdot \sqrt{\frac{C_{\rm vlm}}{N+M}} \cdot \log(1+\frac{N+M}{C_{\rm vlm}}).
\end{flalign}
}
This completes the proof of Theorem~\ref{thm:main}
\end{proof}

%% file: tables/app_examples.tex
\begingroup % Start local scope
\renewcommand{\arraystretch}{1.3} % Adjust row spacing globally

\begin{table}[h!]
\centering
\begin{tabular}{|m{4cm}|m{5cm}|m{5cm}|}
\hline
\includegraphics[width=\linewidth]{figures/examples/164969000.png} 
& \centering\arraybackslash \textbf{Q:}  What green veggie is on the pizza 
& \centering\arraybackslash \textbf{GT:} pepper \\ \hline
\textbf{Match:} & \multicolumn{2}{m{10cm}|}{The pizza has green pepper slices on one of its sections.} \\ \hline
\textbf{Corruption:} & \multicolumn{2}{m{10cm}|}{The pizza has green broccoli florets on one of its sections.} \\ \hline
\textbf{Irrelevance:} & \multicolumn{2}{m{10cm}|}{Beckham obtained his early education at Roseland Academy in Bardstown.
 In 1881 he served as a page in the Kentucky House of Representatives at the age of 12. 
Later, he enrolled at Central University ( now Eastern Kentucky University ) in Richmond, Kentucky but was forced to quit school at the age of 17 to support his widowed mother. 
Two years later, he became principal of Bardstown public schools, serving from 1888 to 1893. Concurrently, he studied law at the University of Kentucky, where he earned his law degree in 1889. 
He was admitted to the bar and commenced practice in Bardstown in 1893. 
He also served as president of the Young Democrats ' Club of Nelson County .} \\ \hline
\end{tabular}
\caption{Illustration of matching, corrupted, and irrelevant information in a sample from VQAv2.}
\label{table:example_vqav2}
\end{table}

\begin{table}[h!]
\centering
\begin{tabular}{|m{4cm}|m{6cm}|m{6cm}|}
\hline
\includegraphics[width=\linewidth]{figures/examples/docvqa_49177.png} 
& \centering\arraybackslash \textbf{Q:}  What time is ‘question and answers ‘session?
& \centering\arraybackslash \textbf{GT:} 12:25 to 12:58 p.m. \\ \hline
\textbf{Match:} & \multicolumn{2}{m{12cm}|}{The 'Questions and Answers' session is scheduled from 12:25 to 12:58 p.m.} \\ \hline
\textbf{Corruption:} & \multicolumn{2}{m{12cm}|}{The 'Questions and Answers' session is scheduled from 2:00 to 5:00 p.m.} \\ \hline
\textbf{Irrelevance:} & \multicolumn{2}{m{12cm}|}{The Americans knew of the approach of the Japanese forces from reports from native scouts and their own patrols , but did not know exactly where or when they would attack . The ridge around which Edson deployed his men consisted of three distinct hillocks . At the southern tip and surrounded on three sides by thick jungle was Hill 80 ( so named because it rose 80 ft ( 24 m ) above sea level ) . Six hundred yards north was Hill 123 ( 123 ft ( 37 m ) high ) , the dominant feature on the ridge . The northernmost hillock was unnamed and about 60 ft ( 18 m ) high . Edson placed the five companies from the Raider battalion on the west side of the ridge and the three Parachute battalion companies on the east side , holding positions in depth from Hill 80 back to Hill 123 . Two of the five Raider companies , \" B \" and \" C \" , held a line between the ridge , a small , swampy lagoon , and the Lunga River . Machine @-@ gun teams from \" E \" Company , the heavy weapons company , were scattered throughout the defenses . Edson placed his command post on Hill 123 .} \\ \hline
\end{tabular}
\caption{Illustration of matching, corrupted, and irrelevant information in a sample from DocVQA.}
\label{table:example_docvqa}
\end{table}

\begin{table}[h!]
\centering
\begin{tabular}{|m{4cm}|m{6cm}|m{6cm}|}
\hline
\includegraphics[width=\linewidth]{figures/examples/mathvista_img2.png} 
& \centering\arraybackslash \textbf{Q:}  Hint: Please answer the question requiring an integer answer and provide the final value, e.g., 1, 2, 3, at the end.
Question: what is the total volume of the measuring cup? (Unit: g)
& \centering\arraybackslash \textbf{GT:} 1000 \\ \hline
\textbf{Match:} & \multicolumn{2}{m{13cm}|}{The measuring cup has markings up to 1000 grams, indicating its total volume capacity.} \\ \hline
\textbf{Corruption:} & \multicolumn{2}{m{13cm}|}{The measuring cup has markings up to 500 grams, indicating its total volume capacity.} \\ \hline
\textbf{Irrelevance:} & \multicolumn{2}{m{13cm}|}{The windmill at Thelnetham was built by millwright George Bloomfield for William Button in 1819 . It replaced a post mill which had been moved to Sandy Lane , Diss , Norfolk the previous year . The mill was set to work on Christmas Day 1819 . In 1832 , the mill was modernised by the fitting of a cast @-@ iron windshaft , Patent sails and a fantail . The new windshaft was fitted on 16 July 1832 . It was made by J Aickman , the Kings Lynn millwright , and weighs 1 \u00be tons ( 1 @,@ 780 kg ) . A new stock was fitted in September 1836 . William Button died on 11 February 1837 . The mill passed jointly to his widow Rebecca and their son Richard . Richard Button worked the mill until 1860 , at which date it was conveyed to his sons Richard and William , who sold it to Richard Peverett from Kenninghall , Norfolk in 1862. } \\ \hline
\end{tabular}
\caption{Illustration of matching, corrupted, and irrelevant information in a sample from MathVista.}
\label{table:example_mathvista}
\end{table}

\begin{table}[h!]
\centering
\begin{tabular}{|m{6cm}|m{6cm}|m{4cm}|}
\hline
\includegraphics[width=\linewidth]{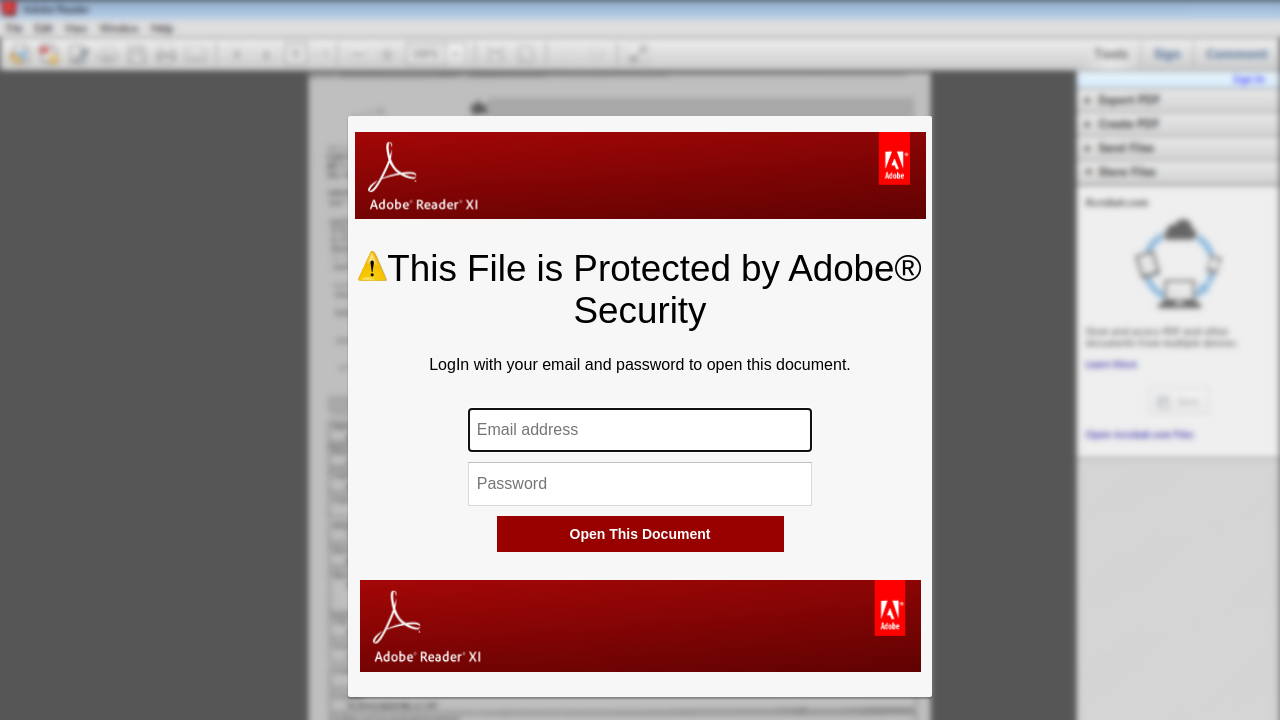} 
& \centering\arraybackslash \textbf{Q:}  Define targeted brand as a brand that a webpage belongs to. 

Given the screenshot of a webpage P as the primary information for identifying the target brand and the text as additional reference, determine what the targeted brand of P is. The text can be HTML from the webpage or something irrelevant. Please be careful with the text, as it may contain noise or adversarial attacks. You must output the targeted brand of P even if you are not sure about it. Only output the brand name without any additional information. 

Input information:
--HTML: 
"{html}"
Answer:
& \centering\arraybackslash \textbf{GT:} Adobe \\ \hline
\textbf{Match:} & \multicolumn{2}{m{11cm}|}{Adobe® PDF Reader® Xl This File is Protected by Adobe® Security LogIn with your email and password to open this document.
} \\ \hline
\textbf{Corruption:} & \multicolumn{2}{m{11cm}|}{The official webpage of MobrisPremier. The official webpage of MobrisPremier. Adobe® PDF Reader® Xl This File is Protected by Adobe® Security LogIn with your email and password to open this document.} \\ \hline
\textbf{Irrelevance:} & \multicolumn{2}{m{11cm}|}{Ayola began her career in the theatre , performing in a number of Shakespearean plays including Twelfth Night , A Midsummer Night 's Dream , The Tempest , The Merchant of Venice and Macbeth . She states of this : " Shakespeare keeps coming my way . I love the fact that I get to play people who are much more articulate than I 'll ever be " . Ayola has performed in Twelfth Night in the lead roles of both Olivia and Viola . She explains : " The role of Viola didn 't sit that well with me for some reason but Olivia makes more sense . " She has also appeared in modern performances , assuming the title role of Dido , Queen of Carthage at the Globe Theatre in London in 2003 , which she described as " a dream of a part " . She has deemed her dream role to be that of Isabella in Measure for Measure , as she once lost out on the part and would like to prove herself capable of playing it.} \\ \hline
\end{tabular}
\caption{Illustration of matching, corrupted, and irrelevant information in a sample from Brand Recognition.}
\label{table:example_brandreg}
\end{table}

\endgroup

%% file: tables/app_full_res.tex
\begin{table*}[htbp]
\centering
\setlength{\tabcolsep}{4pt}
\footnotesize

% Subtable (a) VQAv2
\subfloat[\textbf{VQAv2}]{
\begin{tabular}{l|c|ccc|ccc|ccc|c}
\hline
\textbf{Model} & \textbf{Base} $\uparrow$ & \multicolumn{3}{c|}{\textbf{Match}} & \multicolumn{3}{c|}{\textbf{Corruption}} & \multicolumn{3}{c|}{\textbf{Irrelevance}} & \textbf{Macro} $\uparrow$ \\
\cline{3-11}
& & Accuracy $\uparrow$ & Norm $\uparrow$ & TPR & Accuracy $\uparrow$ & Norm $\uparrow$ & TPR $\downarrow$ & Accuracy $\uparrow$ & Norm $\uparrow$ & TPR $\downarrow$ &  \\
\hline
GPT-4o mini & 69.82 & 87.49 & \underline{125.31} & 89.15 & 51.55 & 73.83 & 52.42 & 72.11 & \underline{103.28} & 3.77 & 70.38 \\
Claude Haiku & 51.02 & 82.81 & \textbf{162.31} & 86.74 & 26.33 & 51.61 & 82.71 & 51.10 & 100.16 & 13.95 & 53.41 \\
GPT-4o & 78.39 & 89.27 & 113.88 & 69.03 & \textbf{70.75} & \underline{90.25} & \underline{27.09} & 78.82 & 100.55 & 1.56 & \textbf{79.61} \\
Claude Sonnet & 66.88 & 77.85 & 116.40 & 49.86& \underline{68.17} & \textbf{101.93} & \textbf{9.58} & 70.89 & \textbf{106.00} & \underline{1.38} & 72.30 \\
\hdashline[0.5pt/2pt]
LLaVA-NeXT-7B & 79.45 & 92.32 & 116.20 & 86.25 & 28.69 & 36.11 & 85.52 & 79.43 & 99.97 & 4.72 & 66.81 \\
LLaVA-NeXT-13B & 81.02 & \textbf{93.59} & 115.51 & 86.45 & 37.61 & 46.42 & 74.43 & \underline{81.29} & 100.33 & 3.30 & 70.83 \\
LLaVA-NeXT-34B & \underline{82.96} & \underline{93.07} & 112.19 & 79.10 & 42.87 & 51.68 & 67.56 & 79.64 & 95.99 & 2.70 & 71.86 \\
Phi3.5 & 75.65 & 91.23 & 120.59 & 79.51 & 35.23 & 46.57 & 74.05 & 74.87 & 98.97 & 2.25 & 67.11 \\
Molmo-7B-D & 76.33 & 88.57 & 116.04 & 88.32 & 49.29 & 64.57 & 59.40 & 76.50 & 100.22 & 9.36 & 71.45 \\
Qwen2-VL-7B & \textbf{85.51} & 92.76 & 108.48 & 13.17& 50.79 & 59.40 & 29.22 & \textbf{83.70} & 97.88 & \textbf{1.28} & \underline{75.75} \\
\hline
\end{tabular}
}

% Subtable (b) DocVQA
\subfloat[\textbf{DocVQA}]{
\begin{tabular}{l|c|ccc|ccc|ccc|c}
\hline
\textbf{Model} & \textbf{Base} $\uparrow$ & \multicolumn{3}{c|}{\textbf{Match}} & \multicolumn{3}{c|}{\textbf{Corruption}} & \multicolumn{3}{c|}{\textbf{Irrelevance}} & \textbf{Macro} $\uparrow$ \\
\cline{3-11}
& & Accuracy $\uparrow$ & Norm $\uparrow$ & TPR & Accuracy $\uparrow$ & Norm $\uparrow$ & TPR $\downarrow$ & Accuracy $\uparrow$ & Norm $\uparrow$ & TPR $\downarrow$ & \\
\hline
GPT-4o mini & 69.40 & 81.40 & 117.26 & 82.74 & 38.20 & 55.04 & 52.07 & 67.20 & 96.83 & 0.80 & 62.27 \\
Claude Haiku & 69.53 & 83.45 & 120.06 & 68.77 & 39.35 & 56.61 & 47.67 & 57.82 & 83.16 & 1.18 & 60.21 \\
GPT-4o & 85.00 & 90.40 & 106.35 & 64.75 & \underline{73.60} & \underline{86.59} & \underline{17.96} & 86.40 & \textbf{101.65} & \underline{0.23} & \underline{83.47} \\
Claude Sonnet & \underline{87.00} & 91.53 & 105.15 & 41.18& \textbf{84.60} & \textbf{97.24} & \textbf{3.21} & \underline{87.41} & 100.47 & \textbf{0.00} & \textbf{87.85} \\
\hdashline[0.5pt/2pt]
LLaVA-NeXT-7B & 53.60 & 90.80 & \textbf{169.40} & 86.92 & 10.00 & 18.66 & 87.77 & 52.40 & 97.76 & 0.71 & 51.07 \\
LLaVA-NeXT-13B & 57.70 & 90.40 & \underline{156.68} & 87.82 & 11.00 & 19.06 & 86.84 & 55.80 & 96.68 & 0.65 & 52.40 \\
LLaVA-NeXT-34B & 64.00 & 87.80 & 137.19 & 84.62 & 15.10 & 23.59 & 82.69 & 62.70 & 97.97 & 0.13 & 55.20 \\
Phi3.5 & 78.20 & \underline{92.40} & 118.16 & 58.01 & 50.50 & 64.60 & 40.51 & 77.00 & 98.46 & \textbf{0.00} & 73.30 \\
Molmo-7B-D & 74.00 & 90.30 & 122.30 & 87.54 & 38.40 & 51.89 & 57.20 & 74.70 & \underline{100.95} & 0.37 & 67.80 \\
Qwen2-VL-7B & \textbf{90.50} & \textbf{95.10} & 105.08 & 51.97& 57.50 & 63.64 & 37.41 & \textbf{89.90} & 99.34 & 0.22 & 80.83 \\
\hline
\end{tabular}
}

% Subtable (c) MathVista
\subfloat[\textbf{MathVista}]{
\begin{tabular}{l|c|ccc|ccc|ccc|c}
\hline
\textbf{Model} & \textbf{Base} $\uparrow$ & \multicolumn{3}{c|}{\textbf{Match}} & \multicolumn{3}{c|}{\textbf{Corruption}} & \multicolumn{3}{c|}{\textbf{Irrelevance}} & \textbf{Macro} $\uparrow$ \\
\cline{3-11}
& & Accuracy $\uparrow$ & Norm $\uparrow$ & TPR & Accuracy $\uparrow$ & Norm $\uparrow$ & TPR $\downarrow$ & Accuracy $\uparrow$ & Norm $\uparrow$ & TPR $\downarrow$ & \\
\hline
GPT-4o mini & 52.30 & 73.80 & 141.11 & 88.82 & 23.90 & 45.70 & 80.28 & 44.40 & 84.89 & 20.14 & 47.37 \\
Claude Haiku & 41.00 & \textbf{80.30} & 195.85 & 88.04 & 19.80 & 48.29 & 77.42 & 39.70 & 96.83 & 23.33 & 46.60 \\
GPT-4o & \textbf{58.90} & 73.70 & 125.04 & 85.20 & \underline{41.20} & 69.95 & \underline{48.98} & 53.10 & 90.15 & 13.55 & \underline{56.00} \\
Claude Sonnet & \underline{56.30} & 68.10 & 120.95 & 57.69& \textbf{49.30} & \textbf{87.57} & \textbf{29.14} & \textbf{55.20} & 98.05 & \textbf{7.96} & \textbf{57.53} \\
\hdashline[0.5pt/2pt]
LLaVA-NeXT-7B & 35.80 & 74.80 & \textbf{273.62} & 88.72 & 19.70 & 54.97 & 84.19 & 28.40 & \textbf{104.02} & 38.22 & 40.97 \\
LLaVA-NeXT-13B & 36.20 & 76.20 & \underline{257.43} & 88.98 & 20.60 & 56.89 & 80.83 & 32.60 & 96.28 & 37.18 & 43.13 \\
LLaVA-NeXT-34B & 34.00 & 68.00 & 200.00 & 73.59& 21.70 & 61.98 & 67.64 & 32.10 & 94.41 & 20.40 & 40.60 \\
Phi3.5 & 43.10 & 73.70 & 171.21 & 84.82 & 22.20 & 51.47 & 80.20 & 41.10 & 95.36 & 13.99 & 45.67 \\
Molmo-7B-D & 44.90 & 68.50 & 152.57 & 82.46 & 32.90 & \underline{73.27} & 60.63 & 45.30 & \underline{100.89} & 27.49 & 48.90 \\
Qwen2-VL-7B & 55.40 & \underline{77.80} & 140.43 & 84.50 & 28.90 & 52.18 & 70.23 & \underline{54.90} & 99.10 & \underline{8.44} & 53.87 \\
\hline
\end{tabular}
}

% Subtable (d) Brand Detection
\subfloat[\textbf{Brand Detection}]{
\begin{tabular}{l|c|ccc|ccc|ccc|c}
\hline
\textbf{Model} & \textbf{Base} $\uparrow$ & \multicolumn{3}{c|}{\textbf{Match}} & \multicolumn{3}{c|}{\textbf{Corruption}} & \multicolumn{3}{c|}{\textbf{Irrelevance}} & \textbf{Macro} $\uparrow$ \\
\cline{3-11}
& & Accuracy $\uparrow$ & Norm $\uparrow$ & TPR & Accuracy $\uparrow$ & Norm $\uparrow$ & TPR $\downarrow$ & Accuracy $\uparrow$ & Norm $\uparrow$ & TPR $\downarrow$ & \\
\hline
GPT-4o mini & 88.84 & 86.88 & 97.80 & 30.43 & 84.80 & 95.44 & 7.48 & 88.48 & 99.60 & 0.08 & 86.72 \\
Claude Haiku & 84.40 & 83.40 & 98.81 & 26.02 & 78.72 & 93.27 & 6.44 & 82.28 & 97.49 & \textbf{0.00} & 81.47 \\
GPT-4o & 88.68 & \underline{89.48} & \underline{100.90} & 14.64& \underline{89.76} & \textbf{101.22} & \textbf{0.83} & \underline{89.16} & \textbf{100.54} & \underline{0.04} & \underline{89.47} \\
Claude Sonnet & \textbf{90.20} & \textbf{90.56} & 100.40 & 17.03& \textbf{90.24} & \underline{100.04} & \underline{0.96} & \textbf{90.24} & \underline{100.04} & \textbf{0.00} & \textbf{90.35} \\
\hdashline[0.5pt/2pt]
LLaVA-NeXT-7B & 78.60 & 77.56 & 98.67 & 82.39 & 62.52 & 79.54 & 64.74 & 16.28 & 20.72 & 70.45 & 52.12 \\
LLaVA-NeXT-13B & 83.00 & 79.00 & 95.18 & 77.04 & 33.96 & 40.92 & 72.97 & 11.72 & 14.12 & 79.61 & 41.56 \\
LLaVA-NeXT-34B & 66.28 & 68.28 & \textbf{102.99} & 31.60 & 53.52 & 80.77 & 23.49 & 52.84 & 79.69 & 10.65 & 58.21 \\
Phi3.5 & 84.40 & 83.84 & 99.33 & 31.39 & 60.68 & 71.90 & 50.54 & 16.44 & 19.48 & 79.17 & 53.65 \\
Molmo-7B-D & 87.44 & 87.32 & 99.86 & 37.38 & 41.44 & 47.39 & 60.40 & 60.88 & 69.63 & 27.36 & 63.21 \\
Qwen2-VL-7B & \underline{89.68} & 88.92 & 99.15 & 17.22 & 86.48 & 96.43 & 2.99 & 70.16 & 78.20 & 15.73 & 81.85 \\
\hline
\end{tabular}
}
\caption{Performance in Accuracy, Normalized Accuracy (Norm) and Text Preference Ratio (TPR) across four datasets under three text variations: Match, Corruption, and Irrelevance. The \textbf{Macro} column represents the average of Match, Corruption, and Irrelevance \textbf{Accuracy} for each model, calculated to be comparable to the \textbf{Base} accuracy.}
\label{app:table:full_res}
\end{table*}

%% file: tables/app_ft_res_generalization.tex
\begin{table*}[htbp]
\centering
\setlength{\tabcolsep}{4pt}
\begingroup % Start local scope
\renewcommand{\arraystretch}{1.2} % Adjust row spacing globally

\footnotesize

\subfloat[\textbf{VQAv2}]{
\begin{tabular}{l|c|ccc|ccc|ccc|c}
\hline
\textbf{Model}                   & \textbf{Base} $\uparrow$ & \multicolumn{3}{c|}{\textbf{Match}}          & \multicolumn{3}{c|}{\textbf{Corruption}}      & \multicolumn{3}{c|}{\textbf{Irrelevance}}     & \textbf{Macro} $\uparrow$ \\
\cline{3-11}
                                 &                          & Accuracy $\uparrow$ & Norm $\uparrow$ & TPR & Accuracy $\uparrow$ & Norm $\uparrow$ & TPR & Accuracy $\uparrow$ & Norm $\uparrow$ &   TPR &                       \\
\hline
LLaVA-NeXT-7B                    & \textbf{79.45}                    & 92.32               & 116.20          & 86.25& 28.69               & 36.11          & 85.52         & 79.43               & 99.97          & 4.72           & 66.81                    \\
Instruction                      & \textbf{79.45}                    & 92.25               & 116.12          & 86.46& 34.27               & 43.13          & 78.50& 78.15               & 98.36          & 6.69& 68.22                    \\
SFT                              & 77.48                    & 87.56               & 113.01          & 59.73         & 71.25               & 91.94          & 20.00         & 77.32               & 99.79          & 4.06           & \textbf{78.71}                    \\
\hline
Qwen2-VL-7B                      & \textbf{85.51}                    & 92.76               & 108.48          & 13.17         & 50.79               & 59.40          & 29.22         & 83.70               & 97.88          & 1.28           & 75.75                    \\
Instruction                      & \textbf{85.51}                    & 92.62               & 108.32          & 14.42& 54.78               & 64.07          & 27.01& 82.82               & 96.85          & 1.18& 76.74                    \\
SFT                              & 84.18                    & 87.01               & 103.36          & 36.65         & 82.72               & 98.26          & 6.69          & 84.00               & 99.79          & 2.59           & \textbf{84.58 }                   \\
\hline
\end{tabular}
}

\subfloat[\textbf{DocVQA}]{
\begin{tabular}{l|c|ccc|ccc|ccc|c}
\hline
\textbf{Model}                   & \textbf{Base} $\uparrow$ & \multicolumn{3}{c|}{\textbf{Match}}          & \multicolumn{3}{c|}{\textbf{Corruption}}      & \multicolumn{3}{c|}{\textbf{Irrelevance}}     & \textbf{Macro} $\uparrow$ \\
\cline{3-11}
                                 &                          & Accuracy $\uparrow$ & Norm $\uparrow$ & TPR & Accuracy $\uparrow$ & Norm $\uparrow$ & TPR & Accuracy $\uparrow$ & Norm $\uparrow$ & TPR &                                             \\
\hline
LLaVA-NeXT-7B                    & \textbf{53.60}                    & 90.80               & 169.40          & 86.92          & 10.00               & 18.66          & 87.77          & 52.40               & 97.76          & 0.71          & 51.07                    \\
Instruction                      & \textbf{53.60}                    & 88.60               & 165.30          & 84.01& 9.80                & 18.28          & 87.38& 49.40               & 92.16          & 1.54& 49.27                    \\
SFT                              & 52.20                    & 75.50               & 144.63          & 56.21          & 42.80               & 81.99          & 28.19          & 50.20               & 96.17          & 0.14          & \textbf{56.17}                    \\
\hline
Qwen2-VL-7B                      & \textbf{90.50}                    & 95.10               & 105.08          & 51.97          & 57.50               & 63.64          & 37.41          & 89.90               & 99.34          & 0.22          & 80.83                    \\
Instruction                      & \textbf{90.50}                    & 94.70               & 104.64          & 51.46& 57.80               & 63.88          & 37.00& 89.80               & 99.23          & 0.11& 80.77                    \\
SFT                              & 90.30                    & 93.10               & 103.10          & 26.06          & 84.30               & 93.35          & 6.32           & 89.50               & 99.11          & 0.11          & \textbf{88.97}                    \\
\hline
\end{tabular}
}

\subfloat[\textbf{MathVista}]{
\begin{tabular}{l|c|ccc|ccc|ccc|c}
\hline
\textbf{Model}                   & \textbf{Base} $\uparrow$ & \multicolumn{3}{c|}{\textbf{Match}}          & \multicolumn{3}{c|}{\textbf{Corruption}}      & \multicolumn{3}{c|}{\textbf{Irrelevance}}     & \textbf{Macro} $\uparrow$ \\
\cline{3-11}
                                 &                          & Accuracy $\uparrow$ & Norm $\uparrow$ & TPR  & Accuracy $\uparrow$ & Norm $\uparrow$ & TPR  & Accuracy $\uparrow$ & Norm $\uparrow$ & TPR  &                          \\
\hline
LLaVA-NeXT-7B                    & \textbf{35.80}           & 74.80               & 208.94          & 84.32          & 19.70               & 55.03           & 84.19          & 28.40               & 79.33           & 34.57          & 41.03                    \\
Instruction                      & \textbf{35.80}           & 70.60               & 197.77          & 84.68              & 21.80               & 60.89           & 81.85              & 31.20               & 87.15           & 32.94              & 41.20                    \\
SFT                              & 35.30                    & 68.70               & 194.90          & 77.42          & 23.50               & 66.57           & 63.75          & 32.70               & 92.64           & 10.76          & \textbf{41.63}           \\
\hline
Qwen2-VL-7B                      & 55.40                    & 77.80               & 140.43          & 84.50          & 28.90               & 52.17           & 70.23          & 54.90               & 99.10           & 8.44           & 53.87                    \\
Instruction                      & 55.40                    & 78.10               & 140.79          & 86.50              & 29.30               & 52.88           & 70.59              & 54.90               & 99.10           & 8.11              & 54.10                    \\
SFT                              & \textbf{58.50}           & 74.00               & 126.50          & 78.31          & 40.30               & 68.89           & 49.16          & 57.20               & 97.78           & 5.65           & \textbf{57.17}           \\
\hline
\end{tabular}
}

\subfloat[\textbf{Brand Detection}]{
\begin{tabular}{l|c|ccc|ccc|ccc|c}
\hline
\textbf{Model}                   & \textbf{Base} $\uparrow$ & \multicolumn{3}{c|}{\textbf{Match}}          & \multicolumn{3}{c|}{\textbf{Corruption}}      & \multicolumn{3}{c|}{\textbf{Irrelevance}}     & \textbf{Macro} $\uparrow$ \\
\cline{3-11}
                                 &                          & Accuracy $\uparrow$ & Norm $\uparrow$ & TPR  & Accuracy $\uparrow$ & Norm $\uparrow$ & TPR  & Accuracy $\uparrow$ & Norm $\uparrow$ & TPR  &                          \\
\hline
LLaVA-NeXT-7B                    & 78.60                    & 77.56               & 98.67           & 68.30          & 62.52               & 79.54           & 59.17          & 16.28               & 20.72           & 89.14          & 46.44                    \\
Instruction                      & 78.60                    & 78.36               & 99.70           & 66.57              & 54.84               & 69.77           & 59.63              & 8.88                & 11.30           & 85.26              & 47.36                    \\
SFT                              & \textbf{81.36}           & 78.32               & 96.26           & 37.18          & 69.48               & 85.39           & 17.92          & 69.08               & 84.92           & 9.08           & \textbf{72.29}           \\
\hline
Qwen2-VL-7B                      & \textbf{89.68}           & 88.92               & 99.15           & 17.22          & 86.48               & 96.43           & 2.99           & 70.16               & 78.20           & 15.73          & 81.85                    \\
Instruction                      & \textbf{89.68}           & 88.52               & 98.71           & 17.50              & 87.12               & 97.15           & 1.94              & 77.80               & 86.77           & 9.34              & 84.48                    \\
SFT                              & 89.44                    & 90.08               & 100.72          & 20.32          & 88.76               & 99.24           & 1.43           & 87.40               & 97.72           & 0.71           & \textbf{88.75}           \\
\hline
\end{tabular}
}

\endgroup

\caption{Performance of investigated solutions in Accuracy, Normalized Accuracy (Norm) and Text Preference Ratio (TPR) across four datasets under three text variations: Match, Corruption, and Irrelevance. The \textbf{Macro} column represents the average of Match, Corruption, and Irrelevance \textbf{Accuracy} for each model, calculated to be comparable to the \textbf{Base} accuracy.}
\label{app:tab:ft_res}

\end{table*}